\documentclass[10pt,twocolumn,letterpaper]{article}

\usepackage[pagenumbers]{cvpr}      

\usepackage{graphicx}
\usepackage{amsmath}
\usepackage{amssymb}
\usepackage{changepage}
\usepackage{booktabs}
\usepackage{xcolor}         
\definecolor{ForestGreen}{RGB}{34,139,34}
\usepackage{multirow}
\usepackage{wrapfig}
\usepackage{sidecap}
\usepackage{enumitem}
\usepackage{cuted}
\definecolor{pearDark}{HTML}{2980B9}

\usepackage[pagebackref,breaklinks,colorlinks]{hyperref}
\hypersetup{
    colorlinks=true,
    linkcolor=pearDark,
    filecolor=magenta,      
    urlcolor=blue,
    citecolor=pearDark
}

\usepackage[capitalize]{cleveref}
\crefname{section}{Sec.}{Secs.}
\Crefname{section}{Section}{Sections}
\Crefname{table}{Table}{Tables}
\crefname{table}{Tab.}{Tabs.}

\def\cvprPaperID{3674} 
\def\confName{CVPR}
\def\confYear{2022}

\newcommand{\parag}[1]{\vspace{-3mm}\paragraph{#1}}
\newcommand{\drosophila}{Drosophila melanogaster}

\begin{document}

\title{Overcoming the Domain Gap in Neural Action Representations}

\newcommand*\samethanks[1][\value{footnote}]{\footnotemark[#1]}
\author{
    Semih G\"unel\textsuperscript{1,2}
    \quad
    Florian Aymanns\textsuperscript{2}
    \quad
    Sina Honari\textsuperscript{1}
    \quad
    Pavan Ramdya\textsuperscript{2}\thanks{Equal Contribution}
    \quad
    Pascal Fua\textsuperscript{1}\samethanks \\
    \textsuperscript{1}CVLab, EPFL, {\tt\small \{name.surname\}@epfl.ch}\\
    \textsuperscript{2}Neuroengineering Laboratory, EPFL, {\tt\small \{name.surname\}@epfl.ch} \vspace{2.5mm} 
}
\maketitle

\begin{abstract}

Relating animal behaviors to brain activity is a fundamental goal in neuroscience, with practical applications in building robust brain-machine interfaces. However, the domain gap between individuals is a major issue that prevents the training of general models that work on unlabeled subjects. 

Since 3D pose data can now be reliably extracted from multi-view video sequences without manual intervention, we propose to use it to guide the encoding of neural action representations together with a set of neural and behavioral augmentations exploiting the properties of microscopy imaging. To reduce the domain gap, during training, we swap neural and behavioral data across animals that seem to be performing similar actions. 

To demonstrate this, we test our methods on three very different multimodal datasets; one that features flies and their neural activity, one that contains human neural Electrocorticography (ECoG) data, and lastly the RGB video data of human activities from different viewpoints.

\end{abstract}

\vspace{-5pt}
\section{Introduction}

Neural decoding, the accurate prediction of animal behavior from brain activity, is a fundamental challenge in neuroscience with important applications in the development of robust brain machine interfaces. Recent technological advances have enabled simultaneous recordings of neural activity and behavioral data in experimental animals and humans~\cite{dombeck,Seelig,chen2018imaging,lfads,Ecker2010c,wirelesshuman,largescale}. Nevertheless, our understanding of the complex relationship between behavior and neural activity remains limited.

\begin{figure*}[t]
  \centering
  \includegraphics[width=.95\textwidth]{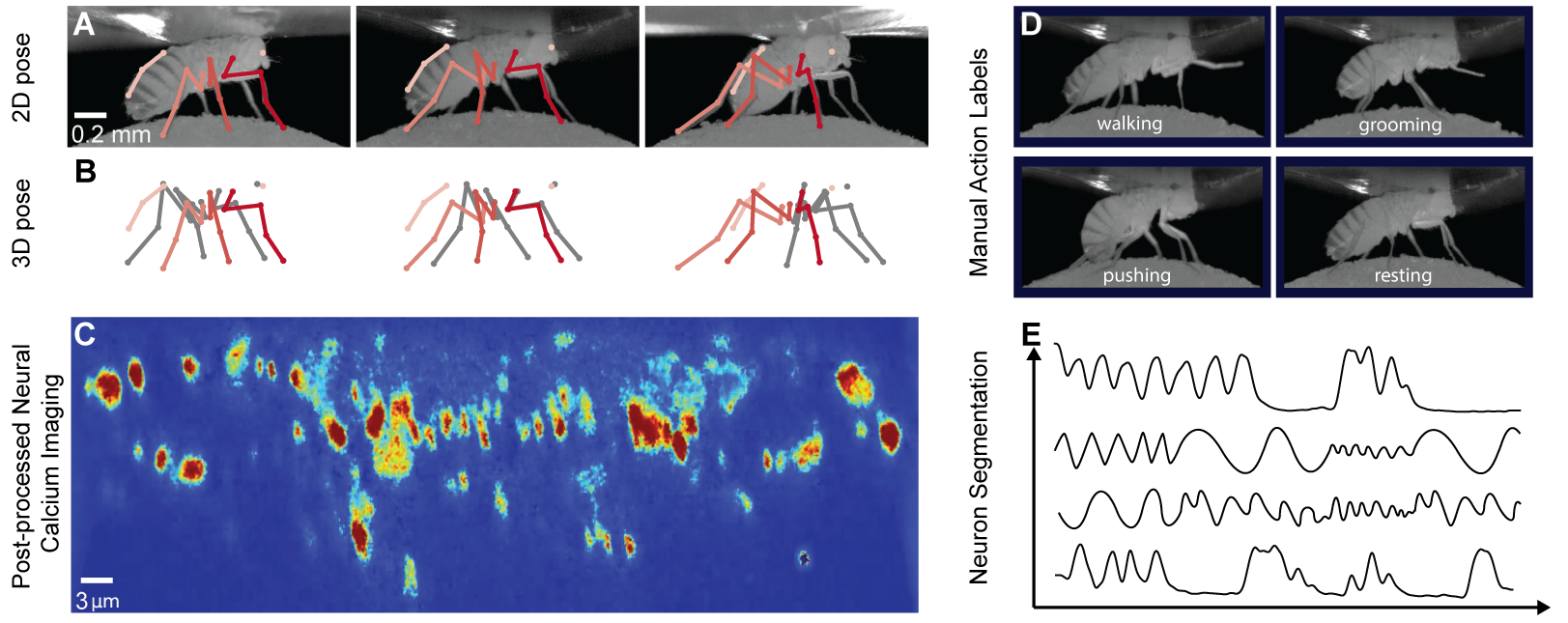}
  \caption{\textbf{Our Motion Capture and Two-Photon (MC2P) Dataset.} A tethered fly (\textit{Drosophila melanogaster}) is recorded using six multi-view infrared cameras and a two-photon microscope. The resulting dataset includes the following. \textbf{(A)} 2D poses extracted from all views (only three are shown), calculated on grayscale images. \textbf{(B)} 3D poses triangulated from the 2D views. \textbf{(C)} Synchronized, registered, and denoised single-channel fluorescence calcium imaging data using a two-photon microscope. Shown are color-coded activity patterns for populations of descending neurons from the brain. These carry action information (red is active, blue is inactive). \textbf{(D)} Annotations for eight animals of eight different behaviors, four of which are shown here. \textbf{(E)} Manual neural segmentation has been performed to extract neural activity traces for each neuron. We will release our MC2P publicly.}
  
  \label{fig:data}
\end{figure*}

\begin{figure*}[t]%
  \centering
  \includegraphics[width=.85\textwidth]%
  {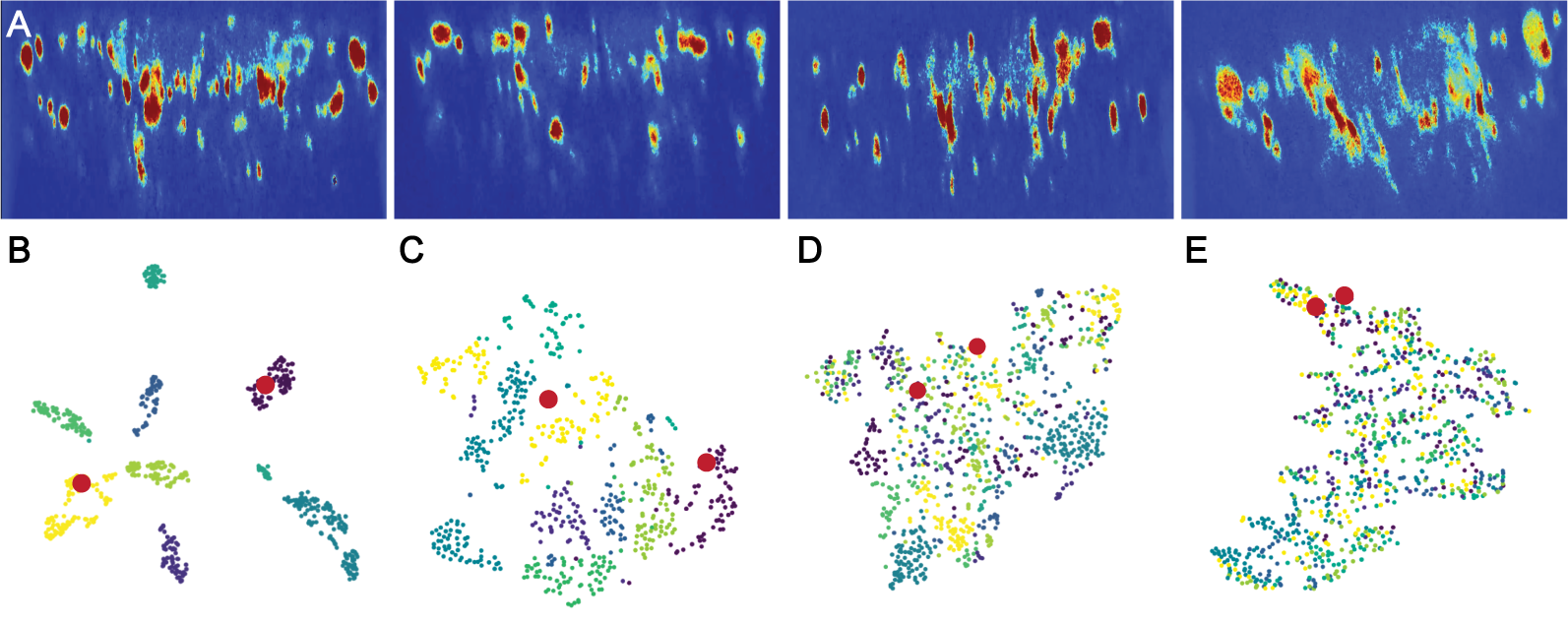}
  \caption{\textbf{Domain gap across nervous systems and learned representations.} \textbf{(Top row)} \textbf{(A)} Neural imaging data from four different animals in our MC2P dataset. Images differ in terms of total brightness, the location and number of visible neurons, and the shape and size of axons. \textbf{(Bottom row)} t-SNE embeddings of different neural representation algorithms on MC2P dataset. Each color denotes a different fly. The two red dots are  embeddings of the same action label for two different animals. t-SNE embeddings of \textbf{(B)} Raw neural data. \textbf{(C)} Contrastive SimCLR~\cite{simclr} representation trained on behavioral and neural pairs, \textbf{(D)} SimCLR and domain adaptation using a two-layer MLP discriminator and a Gradient Reversal Layer \cite{GRL}. \textbf{(E)} Ours. The identify of the animals is discarded and the semantic structure is preserved, as evidenced by the fact that the two red dots are very close to one another.
  }
  \label{fig:domain}
\end{figure*}

A major reason is that it is difficult to obtain many long recordings from mammals and a few subjects are typically not enough to perform meaningful analyses~\cite{pei2021neural}. This is less of a problem when studying the fly \textit{Drosophila melanogaster}, for which long neural and behavioral datasets can be obtained for many individual animals \textbf{(Fig.~\ref{fig:data})}. Nevertheless, current supervised approaches for performing neural decoding ~\cite{supervised1,MLfordecoding} still do not generalize well across animals because each nervous system is unique \textbf{(Fig. \ref{fig:domain}A)}. This creates a significant domain-gap that necessitates tedious and difficult manual labeling of actions. Furthermore, a different model must be trained for each individual animal, requiring more annotation and overwhelming the resources of most laboratories.  

Another problem is that experimental neural imaging data often has unique temporal and spatial properties. The slow decay time of fluorescence signals introduces temporal artifacts. Thus, neural imaging frames include information about an animal's previous behavioral state. This complicates decoding and requires specific handling that standard machine learning algorithms do not provide. 

To address these challenges, we propose to learn neural action representations---embeddings of behavioral states within neural activity patterns---in an unsupervised fashion. To this end, we leverage the recent development of computer vision approaches for automated, markerless 3D pose estimation ~\cite{Gunel19a, dlc} to provide the required supervisory signals without human intervention. We first show that using contrastive learning to generate latent vectors by maximizing the mutual information of simultaneously recorded neural and behavioral data modalities is not sufficient to overcome the domain gap between animals and to generalize to unlabeled animals at test time \textbf{(Fig.~\ref{fig:domain}C)}. To address this problem, we introduce two sets of techniques:
\begin{enumerate}[leftmargin=*]

\item To close the domain gap between animals, we leverage 3D pose information. Specifically, we use pose data to find sequences of similar actions between a source and multiple target animals. Given these sequences, we input to our model neural data from one animal and behavioral data composed of multiple animals. This allows us to train our decoder to ignore animal identity and close the domain gap.

\item To mitigate the slowly decaying calcium data impact from past actions on neural images, we add simulated randomized versions of this effect to our training neural images in the form of a temporally exponentially decaying random action. This trains our decoder to learn the necessary invariance and to ignore the real decay in neural calcium imaging data. Similarly, to make the neural encoders robust to imaging noise resulting from low image spatial resolution, we mix random sequences into sequences of neural data to replicate this noise.

\end{enumerate}
The combination of these techniques allowed us to bridge the domain gap across animals in an unsupervised manner, making it possible to perform action recognition on unlabeled animals \textbf{(Fig.~\ref{fig:domain}E)} better than earlier techniques, including those requiring supervision~\cite{MLfordecoding,behavenet,eegselfsupervised}.

To test the generalization capacity of neural decoding algorithms, we record and use MC2P dataset, which we will make publicly available. It includes two-photon microscope recordings of multiple spontaneously behaving \textit{Drosophila}, and associated behavioral data together with action labels.

Finally, to demonstrate that our technique generalizes beyond this one dataset, we tested it on two additional ones: One that features neural ECoG recordings and 2D pose data for epileptic patients~\cite{peterson_2021,SINGH2021109199} along with the well-known H36M dataset~\cite{Ionescu14a} in which we treat the multiple views as independent domains. Our method markedly improves across-subject action recognition in all datasets.

We hope our work will inspire the use and development of more general unsupervised neural feature extraction algorithms in neuroscience. These approaches promise to accelerate our understanding of how neural dynamics give rise to complex animal behaviors and can enable more robust neural decoding algorithms to be used in brain-machine interfaces.

\vspace{-5pt}
\section{Related Work}

\paragraph{Neural Decoding.} The ability to infer behavioral intentions from neural data, or neural decoding, is essential for the development of effective brain-machine interfaces and for closed-loop experimentation~\cite{bmi,closed-loop}. Neural decoders can be used to increase mobility of patients with disabilities~\cite{COLLINGER201884, GANZER2020763}, or neuromuscular diseases~\cite{neuromusculardisease}, and can expand our understanding of how the nervous system works~\cite{biologicalinsight}. However, most neural decoding methods require manual annotations of training data that are both tedious to acquire and error prone~\cite{MLfordecoding, decodingannotation}. 

Existing self-supervised neural decoding methods~\cite{Wang2018AJILEMP,eegselfsupervised, eegcontrastive, labelingneural} cannot be used on unlabeled subjects without action labels. A potential solution would be to use domain adaptation techniques to treat each new subject as a new domain. However, existing domain adaptation studies of neural decoding~\cite{domainadaptneuraldecoding, stablebrainmachine} have focused on gradual domain shifts associated with slow changes in sensor measurements rather than the challenge of generalizing across individual subjects. In contrast to these methods, our approach is self-supervised and can generalize to unlabeled subjects at test time, without requiring action labels for new individuals.

\paragraph{Action Recognition.}

Contrastive learning has been extensively used on human motion sequences to perform action recognition using 3D pose data~\cite{liu2020snce, su2020predict, Lin_2020} and video-based action understanding~\cite{pan2021videomoco, Dave2021TCLRTC}. However, a barrier to using these tools in neuroscience is that the statistics of our neural data---the locations and sizes of cells---and behavioral data---body part lengths and limb ranges of motion---can be very different from animal to animal, creating a large domain gap.

In theory, there are multimodal domain adaptation methods for action recognition that could deal with this gap~\cite{munro20multi, chen2019temporal, xu2021aligning}. However, they assume supervision in the form of labeled source data. In most laboratory settings, where large amounts data are collected and resources are limited, this is an impractical solution.

\paragraph{Representation Learning.} 

Most efforts to derive a low dimensional representation of neural activity have used recurrent models \cite{Nassar2019TreeStructuredRS, Linderman621540, pmlr-v54-linderman17a}, variational autoencoders \cite{Gao2016LinearDN, lfads}, and dynamical systems~\cite{Multiscale, neuralrecordingtech}. To capture low-dimensional behavioral information, recent methods have enabled markerless predictions of 2D \cite{sleap, graphpose, Bala, couzin, li2020deformation} and 3D poses in animals ~\cite{anipose, Gunel19a, lp3d, zebradataset}. Video and pose data have previously been used to segment and cluster temporally related behavioral information \cite{task_programming, Segalin2020, berman21, quantify, robertZebraFish20}. 

By contrast, there have been relatively few approaches developed to extract behavioral information from neural imaging data~\cite{behavenet, subspace, MLfordecoding}. Most have focused on identifying linear relationships between these two modalities using simple supervised methods, such as correlation analysis, generalized linear models \cite{subtrate, musall19, stringer19}, or regressive methods \cite{behavenet}. 
We present the first joint modeling of behavioral and neural modalities to fully extract behavioral information from neural data using a self-supervised learning technique.

\section{Approach}
Our goal is to be able to interpret neural data such that, given a neural image, one can generate latent representations that are useful for downstream tasks. This is challenging, as there is a wide domain-gap in neural representations for different animals \textbf{(Fig. \ref{fig:domain}A)}. We aimed to leverage unsupervised learning techniques to obtain rich features that, once trained, could be used on downstream tasks including action recognition to predict the behaviors of unlabeled animals. 

Our data is composed of two-photon microscopy neural images synchronized with 3D behavioral data, where we do not know where each action starts and ends. We leveraged contrastive learning to generate latent vectors from both modalities such that their mutual information would be maximized and therefore describe the same underlying action. However, this is insufficient to address the domain-gap between animals \textbf{(Fig. \ref{fig:domain}C)}. To address this issue, we perform swapping augmentation: we replace the original pose or neural data of an animal with one from another set of animals for which there is a high degree of 3D pose similarity at each given instance in time. 

Unlike behavioral data, neural data has unique properties. Neural calcium data contains information about previous actions because it decays slowly across time and it involves limited spatial resolution. To teach our model the invariance of these artifacts of neural data, we propose two data augmentation techniques: (i) Neural Calcium augmentation - given a sequence of neural data, we apply an exponentially decaying neural snapshot to the sequence, which imitates the decaying impact of previous actions, (ii) Neural Mix augmentation - to make the model more robust to noise, we applied a mixing augmentation which merges a sequence of neural data with another randomly sampled neural sequence using a mixing coefficient.

Together, these augmentations enable a self-supervised approach to (i) bridge the domain gap between animals allowing testing on unlabeled ones, and (ii) imitate the temporal and spatial properties of neural data, diversifying it and making it more robust to noise. In the following section, we describe steps in more detail.

\subsection{Problem Definition}

We assume a paired set of data $\mathcal{D}_{s}=\left\{\left(\mathbf{b}_{\mathbf{i}}^{s}, \mathbf{n}_{\mathbf{i}}^{s}\right)\right\}_{i=1}^{n_{s}}$, where $\mathbf{b}^{s}_\mathbf{i}$ and $\mathbf{n}^{s}_\mathbf{i}$ represent behavioral and neural information respectively, with $n_s$ being the number of samples for animal $s\in \mathcal{S}$. 
We quantify behavioral information $\mathbf{b}^{s}_\mathbf{i}$ as a set of 3D poses $\mathbf{b}^{s}_k$ for each frame $k\in\mathbf{i}$ taken of animal $s$, and neural information $\mathbf{n}^{s}_\mathbf{i}$ as a set of two-photon microscope images $\mathbf{n}^{s}_k$, for all frames $ k \in \mathbf{i}$ capturing the activity of neurons. The data is captured such that the two modalities are always synchronized (paired) without human intervention, and therefore describe the same set of events. Our goal is to learn an unsupervised parameterized image encoder function $f_n$, that maps a set of neural images $\mathbf{n}^{s}_\mathbf{i}$ to a low-dimensional representation. We aim for our learned representation to be representative of the underlying action label, while being agnostic to both modality and the identity. We assume that we are not given action labels during unsupervised training. Also note that we do not know at which point in the captured data an action starts and ends. We just have a series of unknown actions performed by different animals.

\subsection{Contrastive Representation Learning}

\begin{figure}[t!]
  \centering
      \includegraphics[width=.50\textwidth]{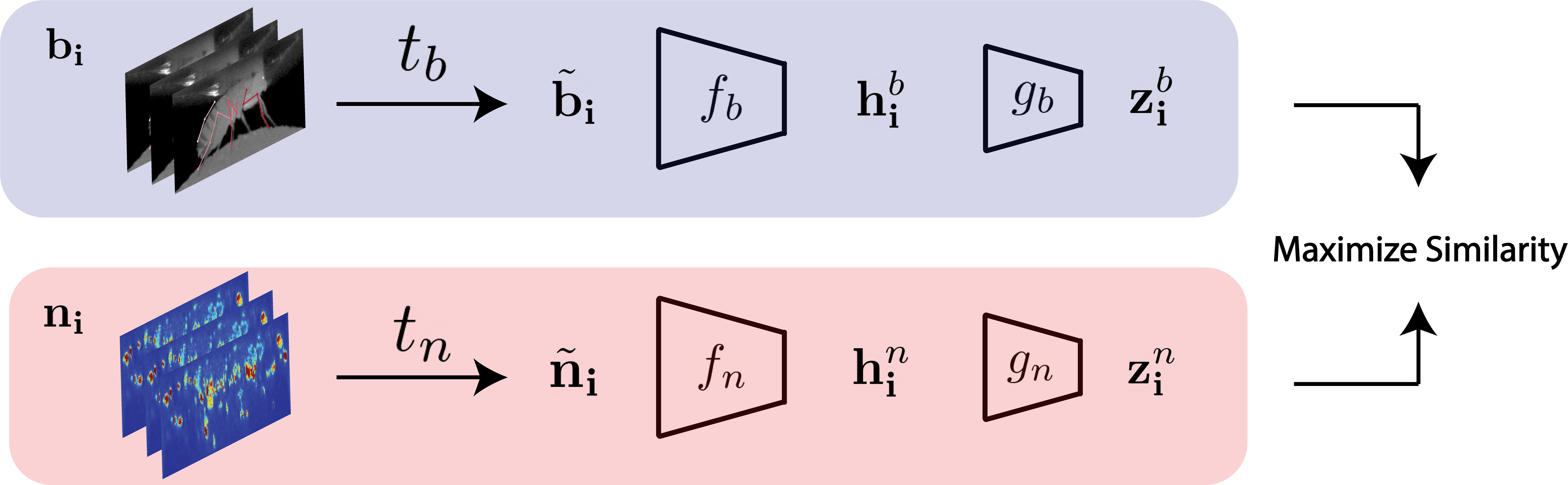}
  \caption{\textbf{Overview of our contrastive learning-based neural action representation learning approach.} First, we sample a synchronized set of behavioral and neural frames, $(\mathbf{b}_{\mathbf{i}}, \mathbf{n}_\mathbf{i})$. Then, we augment these data using randomly sampled augmentation functions $t_b$ and $t_n$. Encoders $f_b$ and $f_n$ then generate intermediate representations $\mathbf{h}^b$ and $\mathbf{h}^n$, which are then projected into $\mathbf{z}_b$ and $\mathbf{z}_t$ by two separate projection heads $g_b$ and $g_n$. We maximize the similarity between the two projections using an InfoNCE loss. At test time,  the red branch and $\textbf{h}_{\textbf{i}}^{n}$ is used for neural decoding.}
  \label{fig:method}
\end{figure}

For each input pair $\left(\mathbf{b}_{\mathbf{i}}^{s}, \mathbf{n}_{\mathbf{i}}^{s}\right)$, we first draw a random augmented version
$( \tilde{\mathbf{b}}_{\mathbf{i}}^{s}, \tilde{\mathbf{n}}_{\mathbf{i}}^{s} ) $ with a sampled transformation function $t_{n} \sim \mathcal{T}_n$ and $t_{b} \sim \mathcal{T}_b$ , where $\mathcal{T}_n$ and $\mathcal{T}_b$ represent a family of stochastic augmentation functions for behavioral and neural data, respectively, which are described in the following sections. Next, the encoder functions $f_b$ and $f_n$ transform the input data into low-dimensional vectors $\mathbf{h}_b$ and $\mathbf{h}_n$, followed by non-linear projection functions $g_b$ and $g_n$, which further transform data into the vectors $\mathbf{z}_b$ and $\mathbf{z}_n$. During training, we sample a minibatch of N input pairs $\left(\mathbf{b}_{\mathbf{i}}^{s}, \mathbf{n}_{\mathbf{i}}^{s}\right)$, and train with the loss function
\begin{equation}
\mathcal{L}^{b\rightarrow n}_{NCE} = - \sum_{i=1}^{N} \log \frac{\exp \left(\left\langle\mathbf{z}^{i}_{b}, \mathbf{z}^{i}_{n}\right\rangle / \tau\right)}{\sum_{k=1}^{N} \exp \left(\left\langle\mathbf{z}^{i}_{b}, \mathbf{z}^{k}_{n}\right\rangle / \tau\right)}
\label{eq:nce}
\end{equation}
where $\left\langle\mathbf{z}^{i}_{b}, \mathbf{z}^{i}_{n}\right\rangle$ is the cosine similarity between behavioral and neural modalities and $\tau \in \mathbb{R}^{+}$ is the temperature parameter. An overview of our method for learning $f_n$ is shown in \textbf{Fig~\ref{fig:method}}. Intuitively, the loss function measures classification accuracy of a N-class classifier that tries to predict $\mathbf{z}^{i}_{n}$ given the true pair $\mathbf{z}^{i}_{b}$.
 To make the loss function symmetric with respect to the negative samples, we also define
 
\begin{equation}
\mathcal{L}^{n\rightarrow b}_{NCE} =  - \sum_{i=1}^{N} \log  \frac{\exp \left(\left\langle \mathbf{z}^{i}_{b}, \mathbf{z}^{i}_{n} \right\rangle / \tau\right)}{\sum_{k=1}^{N} \exp \left(\left\langle \mathbf{z}^{k}_{b}, \mathbf{z}^{i}_{n} \right\rangle / \tau\right)}.
\label{eq:nce2}
\end{equation}

\noindent We take the combined loss function to be $\mathcal{L}_{NCE} = \mathcal{L}^{b\rightarrow n}_{NCE} + \mathcal{L}^{n\rightarrow b}_{NCE}$, as in \cite{zhang2020contrastive, Yuan_2021_CVPR}. The loss function maximizes the mutual information between two modalities \cite{oord2019representation}.  Although standard contrastive learning bridges the gap between different modalities, it does not bridge the gap between different animals \textbf{(Fig.~\ref{fig:domain}C)}. This is a fundamental challenge that we address in this work through augmentations described in the following section, which are part of the neural and behavioral family of augmentations $\mathcal{T}_n$ and $\mathcal{T}_b$. 

\parag{Swapping Augmentation.}
Given a set of consecutive 3D poses $\mathbf{b}^{s}_\mathbf{i}$, for each $k\in\mathbf{i}$, we stochastically replace $\mathbf{b}^{s}_k$ with one of its nearest pose neighbors in the set of domains $\mathcal{D}_{\mathcal{S}}$, where $\mathcal{S}$ is the set of all animals. To do this, we first randomly select a domain $\hat{s} \in \mathcal{S}$ and define a probability distribution $\mathbf{P}^{\hat{s}}_{\mathbf{b}^{s}_k}$ over the domain $\mathcal{D}_{\hat{s}}$ with respect to $\mathbf{b}^{s}_k$, 

\begin{equation}
\mathbf{P}^{\hat{s}}_{\mathbf{b}^{s}_k} (\mathbf{b}^{\hat{s}}_l) = 
\frac{\exp ( - \| \mathbf{b}^{\hat{s}}_l - \mathbf{b}^{s}_k \|_{2})}{\sum_{ \mathbf{b}^{\hat{s}}_m \in  \mathcal{D}_{\hat{s}} } \exp ( - \| \mathbf{b}^{\hat{s}}_m - \mathbf{b}^{s}_k \|_{2})}.
\end{equation}

\begin{figure}[t]
  \centering
      \includegraphics[width=8cm]{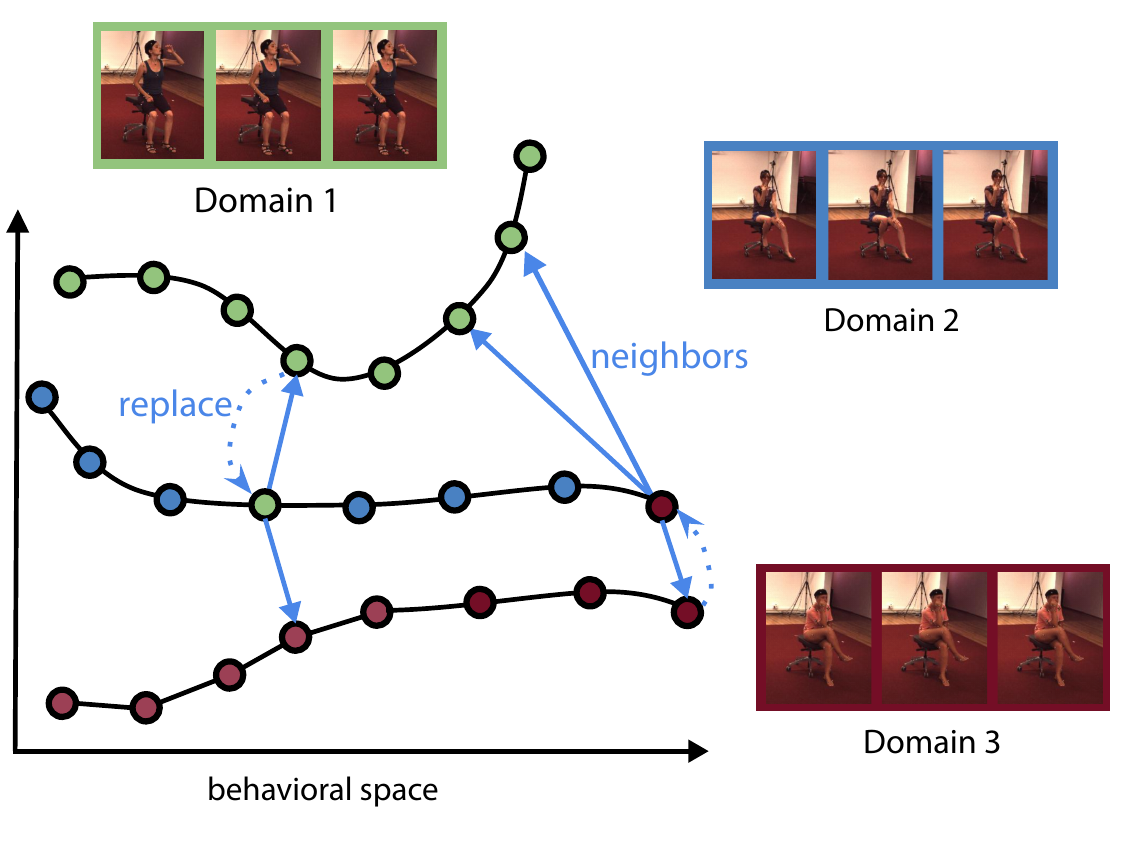}
  \caption{\textbf{Swapping augmentation.} Each 3D pose in the motion sequence of Domain 2 is randomly replaced with its neighbors, from the set of domains $\hat{s} \in \mathcal{S}/s$, which includes Domain 1 and Domain 3. The swapping augmentation hides identity information, while keeping pose changes in the sequence minimal.}
 \label{fig:swap-aug}
\end{figure}

We then replace each 3D pose $\mathbf{b}^{s}_k$ by first uniformly sampling a new domain $\hat{s}$, and then sampling from the above distribution $\mathbf{P}^{\hat{s}}_{\mathbf{b}^{s}_k}$, which yields in $\tilde{\mathbf{b}}^{s}_k \sim \mathbf{P}^{\hat{s}}_{\mathbf{b}^{s}_k}$. In practice, we calculate the distribution $\mathbf{P}$ only over the first $\mathbf{N}$ nearest neighbors of $\mathbf{b}^{s}_k$, in order to sample from a distribution of the most similar poses. We empirically set $\mathbf{N}$ to $128$. Swapping augmentation reduces the identity information in the behavioral data without perturbing it to the extent that semantic action information is lost. Since each behavioral sample $\mathbf{b}^{s}_\mathbf{i}$ is composed of a set of 3D poses, and each 3D pose $\mathbf{b}^{s}_k, \forall k \in \mathbf{i}$ is replaced with a pose of a random domain, the transformed sample $\tilde{\mathbf{b}}_{\mathbf{i}}^{s}$ is now composed of multiple domains. This forces the behavioral encoding function $f_{b}$ to leave identity information out, therefore merging multiple domains in the latent space \textbf{(Fig. \ref{fig:swap-aug})}.

Swapping augmentation is similar to the synonym replacement augmentation used in natural language processing \cite{wei-zou-2019-eda}, where randomly selected words in a sentence are replaced by their synonyms, therefore changing the syntactic form of the sentence without altering the semantics. To the best of our knowledge, we are the first to use swapping augmentation in the context of time-series analysis or for domain adaptation. 

To keep swapping symmetric, we also swap the neural modality. To swap a set of neural images $\mathbf{n}_{\mathbf{i}}^{s}$, we take its behavioral pair $\mathbf{b}_{\mathbf{i}}^{s}$, and searched for similar sets of poses in other domains, with the assumption that similar sets of poses describe the same action. Therefore, once similar behavioral data is found, their neural data can be swapped. Note that, unlike behavior swapping, we do not calculate the distribution on individual 3D pose $\mathbf{b}^{s}_k$, but instead on the whole set of behavioral data $\mathbf{b}_{\mathbf{i}}^{s}$, because similarity in a single pose does not necessarily imply similar actions. More formally, given the behavioral-neural pair $\left(\mathbf{b}_{\mathbf{i}}^{s}, \mathbf{n}_{\mathbf{i}}^{s}\right)$, we swap the neural modality $\mathbf{n}_{\mathbf{i}}^{s}$ with $\mathbf{n}^{\hat{s}}_\mathbf{j}$, with the probability

\begin{equation}
\mathbf{P}^{\hat{s}}_{\mathbf{n}^{s}_\mathbf{i}} (\mathbf{b}^{\hat{s}}_\mathbf{j}) = 
\frac{\exp ( - \| \mathbf{b}^{\hat{s}}_\mathbf{j} - \mathbf{b}^{s}_\mathbf{i} \|_{2})}{\sum_{ \mathbf{b}^{\hat{s}}_\mathbf{m} \in  \mathcal{D}_{\hat{s}} } \exp ( - \| \mathbf{b}^{\hat{s}}_\mathbf{m} - \mathbf{b}^{s}_\mathbf{i} \|_{2})}.
\end{equation}
This results in a new pair $\left(\mathbf{b}_{\mathbf{i}}^{s}, \tilde{\mathbf{n}}_{\mathbf{i}}^{s}\right)$, where the augmented neural data comes from a new animal $\hat{s} \in \mathcal{S}/s$.

\vspace{-5pt}

\parag{Neural Calcium Augmentation.}
Our neural data was obtained using two-photon microscopy and fluorescence calcium imaging. The resulting images are only a function of the underlying neural activity, and have temporal properties that differ from the true neural activity. For example, calcium signals from a neuron change much more slowly than the neuron's actual firing rate. Consequently, a single neural image $\mathbf{n}_t$ includes decaying information concerning neural activity from the recent past, and thus carries information about previous behaviors. This makes it harder to decode the current behavioral state.

We aimed to prevent this overlap of ongoing and previous actions. Specifically, we wanted to teach our network to be invariant with respect to past behavioral information by augmenting the set of possible past actions. 
To do this, we generated new data $\tilde{\mathbf{n}}^{s}_\mathbf{i}$, that included previous neural activity $\mathbf{n}^{s}_k$. To mimic calcium indicator decay dynamics, given a neural data sample $\mathbf{n}^{s}_\mathbf{i}$ of multiple frames, we sample a new neural frame $\mathbf{n}^{s}_k$ from the same domain, where $k \notin \mathbf{i}$. We then convolve $\mathbf{n}^{s}_k$ with the temporally decaying calcium convolutional kernel $\mathcal{K}$, therefore creating a set of images from a single frame $\mathbf{n}^{s}_k$, which we then add back to the original data sample $\mathbf{n}^{s}_\mathbf{i}$. This results in $\tilde{\mathbf{n}}^{s}_\mathbf{i} = \mathbf{n}^{s}_\mathbf{i} + \mathcal{K} * \mathbf{n}^{s}_k$ where $*$ denotes the convolutional operation. 
In the Supplementary Material, we explain calcium dynamics and our calculation of the kernel $\mathcal{K}$ in more detail. 

\parag{Neural Mix Augmentation.} Two-photon microscopy images often include multiple neural signals combined within a single pixel. This is due to the the fact that multiple axons can be present in a small tissue volume that is below the spatial resolution of the microscope. To mimic this noise-adding effect, given a neural image $\mathbf{n}^s_{\textbf{i}}$, we randomly sample a set of frames $\mathbf{n}^{\hat{s}}_{\textbf{k}}$, from a random domain $\hat{s}$. We then return the blend of these two videos,  $\tilde{\mathbf{n}}^s_{\textbf{i}} = \mathbf{n}^s_{\textbf{i}} + \alpha \mathbf{n}^{\hat{s}}_{\textbf{k}}$, to mix and hide the behavioral information. Unlike the CutMix \cite{cutmix} and Mixup \cite{mixup} augmentations used for supervised training, we apply the augmentation in an unsupervised setup to make the model more robust to noise. We sample a random $\alpha$ for the entire set of samples in $\mathbf{n}^s_{\textbf{i}}$.

\section{Experiments}

We test our method on three datasets. In this section, we describe these datasets, the set of baselines against which we compare our model, and finally the quantitative comparison of all models.

\subsection{Datasets}

We ran most of our experiments on a large dataset of fly neural and behavioral recordings that we acquired and describe below. To demonstrate our method's ability to generalize, we also adapted it to run on another multimodal dataset that features neural ECoG recordings and markerless motion capture\cite{peterson_2021,SINGH2021109199}, as well as the well known H36M human motion dataset~\cite{Ionescu14a}.

\parag{MC2P:} Since there was no available neural-behavioral dataset with a rich variety of spontaneous behaviors from multiple individuals, we acquired our own dataset that we name \textit{Motion Capture and Two-photon Dataset (MC2P)}. We will release this dataset publicly.
MC2P features data acquired from tethered behaving adult flies, \textit{Drosophila melanogaster} \textbf{(Fig.~\ref{fig:data})}, It includes:
\begin{enumerate}[leftmargin=*]

    \item Infrared video sequences of the fly acquired using six synchronized and calibrated infrared cameras forming a ring with the animal at its center. The images are $480\times 960$ pixels in size and recorded at $100$ fps. 

    \item Neural activity imaging obtained from the axons of descending neurons that pass from the brain to fly's ventral nerve cord (motor system) and drive actions. The neural images are $480\times 736$ pixels in size and recorded at $16$ fps using a two-photon microscope \cite{chen18} that measures  the calcium influx which is a proxy for the neuron's actual firing rate. 
    
\end{enumerate}
We recorded 40 animals over 364 trials, resulting in 20.7 hours of recordings with 7,480,000 behavioral images and 1,197,025 neural images. We provide additional details and examples in the Supplementary Material. We give an example video of synchronized behavioral and neural modalities in \textbf{Supplementary Videos {\color{pearDark} 1-2}}.

To obtain quantitative behavioral data from video sequences, we extracted 3D poses expressed in terms of the 3D coordinates of 38 keypoints~\cite{Gunel19a}. We provide an example of detected poses and motion capture in \textbf{Supplementary Videos {\color{pearDark} 3-4}}. For validation purposes, we manually annotated a subset of frames using eight behavioral labels: \textit{forward walking}, \textit{pushing}, \textit{hindleg grooming},  \textit{abdominal grooming}, \textit{rest}, \textit{foreleg grooming}, \textit{antenna grooming}, and \textit{eye grooming}.  We provide an example of behavioral annotations in \textbf{Supplementary Video {\color{pearDark} 5}}.

\parag{ECoG dataset~\cite{peterson_2021,SINGH2021109199}:} This dataset was recorded from epilepsy patients over a period of 7-9 days. Each patient had 90 electrodes implanted under their skull. The data comprises human neural  Electrocorticography (ECoG) recordings and markerless motion capture of upper-body 2D poses. The dataset is labeled to indicate periods of voluntary spontaneous motions, or rest. As for two-photon images in flies, ECoG recordings show a significant domain gap across individual subjects. We applied our multi-modal contrastive learning approach on ECoG and 2D pose data along with swapping-augmentation. Then, we applied an across-subject benchmark in which we do action recognition on a new subject without known action labels.

\begin{figure}[t]
  \centering
  \includegraphics[width=.49\textwidth]{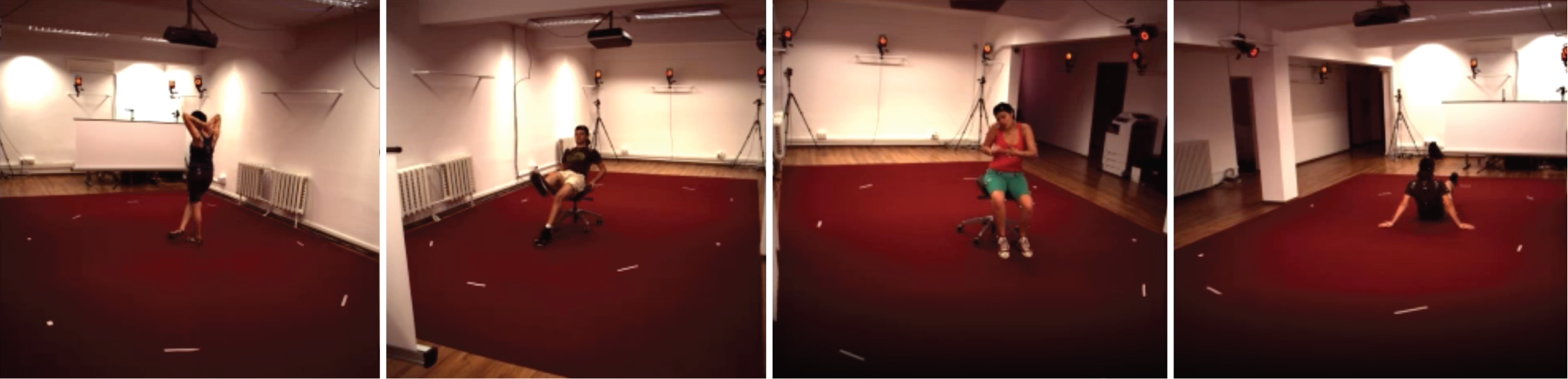}
  \caption{\textbf{Domain Gap in the H3.6M dataset.} Similar to the domain gap across nervous systems, RGB images show a significant domain gap when the camera angle changes across individuals. We guide action recognition across cameras in RGB images using 3D poses and behavioral swapping.}
  \label{fig:h36m_gap}
  \vspace{-10pt}
\end{figure}

\parag{H3.6M \cite{Ionescu14a}:} H3.6M is a multi-view motion capture dataset that is not inherently multimodal. However, to test our approach in a very different context than the other two cases, we treated the videos acquired by different camera angles as belonging to separate domains. Since videos are tied to 3D poses, we used these two modalities and applied swapping augmentation together with multimodal contrastive learning to reduce the domain gap across individuals. Then, we evaluated the learned representations by performing action recognition on a camera angle that we do not have action labels for. This simulates our across-subject benchmark used in the MC2P dataset. For each experiment we selected three actions, which can be classified without examining large window sizes. We give additional details in the Supplementary Material. 

\subsection{Baselines}

\begin{table*}[t]
\begin{minipage}[t]{0.7\textwidth}
\small{
\begin{tabular}{ll|cc|cc|cc}
\toprule
\multicolumn{2}{r|}{\textbf{Tasks $\rightarrow$}} & \multicolumn{2}{c|}{\textbf{Single-Subject $\uparrow$ }} & \multicolumn{2}{c|}{\textbf{Across-Subject $\uparrow$}} &
\multicolumn{2}{c}{\textbf{Identity Recog. $\downarrow$}} \\
\multicolumn{2}{r|}{\textbf{Percentage of Data $\rightarrow$}} & \textbf{0.5} & \textbf{1.0} & \textbf{0.5} & \textbf{1.0} & \textbf{0.5} & \textbf{1.0} \\
\midrule
\multicolumn{2}{l|}{{Random Guess}} &  16.6 & 16.6 & 16.6& 16.6 & 12.5 & 12.5 \\
\midrule
Neural (Linear) & \parbox[t]{1.5mm}{\multirow{2}{*}{\rotatebox[origin=c]{90}{\scriptsize{\textbf{Sup.}}}}} & 29.3  & 32.5 & 18.4 & 18.4 & 100.0 & 100.0 \\
Neural (MLP) & &  -- & --	& 18.4 & 18.4 & 100.0 & 100.0 \\
\midrule
 SeqCLR \cite{eegcontrastive}  &
\parbox[t]{1.5mm}{\multirow{3}{*}{\rotatebox[origin=c]{90}{\scriptsize{\textbf{Self-Supervised}}}}} & 39.5 & 42.1 & 21.9 & 28.4 & 93.0 & 96.5 \\
Ours (Neural Only)  && 42.0 & 44.8 & 21.3 & 30.6 & 94.1 & 96.8 \\
SimCLR\cite{simclr}  && 54.3 & 57.6 & 46.9 & 50.6 & 69.9 & 80.3 \\
Regression (Recurrent) & &  53.6& 59.7 & 49.4 & 51.2 & 89.5 & 91.8\\
Regression (Convolution) & & 52.6 & 59.6 & 50.6 & 55.8 & 88.7 & 92.5\\
BehaveNet \cite{behavenet} & & 54.6 & 60.2 & 50.5 & 56.8 & 80.2 & 83.4\\
Ours & & \textbf{57.9} & \textbf{63.3} & \textbf{54.8} & \textbf{61.9} & \textbf{12.5} & \textbf{12.5}\\
\midrule
SimCLR \cite{simclr} + MMD \cite{MMD} & \parbox[t]{1.5mm}{\multirow{3}{*}{\rotatebox[origin=c]{90}{\scriptsize{\textbf{Domain Ada.}}}}} &  53.6 & 57.8 & 50.1 & 53.1 & 18.4 & 21.2 \\
SimCLR \cite{simclr} + GRL \cite{GRL} &  &  53.5 & 56.3 & 49.9 & 52.3 & 16.7& 19.1 \\
Reg. (Conv.) + MMD \cite{MMD} & & 54.5 & 60.7 & 52.6 & 55.4 & 18.2 & 19.5\\
Reg. (Conv.) + GRL \cite{GRL} & & 55.5 & 60.2 & 51.8 & 55.7 & 17.2 & 17.3 \\
MM-SADA \cite{munro20multi} && 53.1 & 56.2 &  49.2 & 52.1 & 13.8 & 15.2 \\
\bottomrule
\end{tabular}}
\end{minipage}\hfill
\begin{minipage}[t]{0.27\linewidth}
    \vspace{-31mm}
\caption{\textbf{Action Recognition Accuracy.} Single- and Across-Subject action recognition results on the MC2P dataset.  Neural MLP results for the single-subject task are removed because single subject often do not have enough labels for every action. Smaller numbers are better for Identity Recognition. Our method performs better than previous neural decoding methods and other self-supervised learning methods in all benchmarks, while at the same time closing the domain gap between animals, as shown by the identity recognition task.}
\label{tab:accuracy}

\end{minipage}
\end{table*}

We evaluated our method using two supervised baselines, Neural Linear and Neural MLP. These directly predict action labels from neural data without any unsupervised pretraining using cross-entropy loss. We also compared our approach to three regression methods that attempt to regress behavioral data from neural data, which is a common neural decoding technique. These include a recent neural decoding algorithm, BehaveNet~\cite{behavenet}, as well as to two other regression baselines with recurrent and convolutional approaches: Regression (Recurrent) and Regression (Convolution). In addition, we compare our approach to recent self-supervised representation learning methods, including SeqCLR~\cite{eegcontrastive} and SimCLR~\cite{simclr}. We also combine convolutional regression-based method (Reg. (Conv)) or the self-supervised learning algorithm SimCLR with the common domain adaptation techniques Gradient Reversal Layer (GRL)\cite{GRL}, or Mean Maximum Discrepancy \cite{MMD}. This yields four domain adaptation models. Finally, we apply a recent multi-modal domain adaptation network for action recognition, MM-SADA\cite{munro20multi} on MC2P dataset. For all of these methods, we used the same backbone architecture. We describe the backbone architecture and the baseline methods in more detail in the Supplementary Material.

\begin{table}[!h]
\centering
\renewcommand{\arraystretch}{1.1}

{\footnotesize
\begin{tabular}{p{1.1cm}|l|ccc}
\toprule
  & \textbf{Tasks $\rightarrow$}  & \textbf{A.S. $\uparrow$} & \textbf{A.S. $\uparrow$} & \textbf{I.R. $\downarrow$ }\\
 \textbf{Dataset} & \textbf{Percentage of Data $\rightarrow$} & \textbf{0.5} & \textbf{1.0} & \textbf{1.0} \\
\midrule 

{\multirow{4}{\linewidth}{{\scriptsize\textbf{H3.6M Walking, Sitting, Posing}}}} & Random Gu.  &  33.0	& 33.0 & 33.0 \\
\cline{2-5} 
& Supervised &   46.6  &	48.3 &  100.0 \\
\cline{2-5} 
& SimCLR \cite{simclr} (RGB) &  33.2 &	33.5 & 99.5 \\
& SimCLR\cite{simclr} & 53.3 &	55.7 & 99.2 \\
& Regression (Conv.) & 65.2 &68.8 & 68.4 \\
 & Ours & \textbf{72.4} & \textbf{73.6} & \textbf{42.3}\\
 
 \midrule

{\multirow{4}{\linewidth}{{\scriptsize\textbf{H3.6M Walking Together, Directions, Eating}}}} & Random Gu.  &  33.3	& 33.3 & 33.3 \\
\cline{2-5} 
& Supervised &   31.2  &	30.9 &  100.0 \\
\cline{2-5} 
& SimCLR \cite{simclr} (RGB) &   34.6 &	34.4 & 100.0 \\
& SimCLR\cite{simclr} & 52.3 &	53.2 & 94.8 \\
& Regression (Conv.) & 44.8 & 48.7 & 62.1 \\
 & Ours & \textbf{63.2} & \textbf{68.3} & \textbf{44.8}\\
 
 \midrule
 
{\multirow{4}{\linewidth}{{\scriptsize\textbf{ECoG Moving, Rest}}}} & Random Gu.  &  50.0	& 50.0 & 33.3 \\
\cline{2-5} 
& Supervised &   54.2  &	53.8 &  100.0 \\
\cline{2-5} 
& SimCLR \cite{simclr} (ECoG) &   52.3 &	55.1 & 98.0 \\
& SimCLR\cite{simclr} & 64.6 & 72.1 & 81.1 \\
& Regression (Conv.) & 64.1 & 71.8 & 74.3 \\
 & Ours & \textbf{75.8} & \textbf{81.9} & \textbf{53.0}\\

\bottomrule
\end{tabular}
}
\footnotesize{\caption{Across-subject (A.S.) and identity recognition (I.R.) results on H3.6M dataset\cite{h36m_pami} using RGB and 3D pose, and on ECoG Move vs Rest \cite{peterson_2021}  using neural ECoG recordings and 2D pose. For Ours, we remove calcium imaging specific augmentations and only use swapping augmentation. Swapping augmentation closes the domain gap for the contrastive learning and strongly improves across-subject action recognition on both datasets. }\label{tab:results_h36m}}
\end{table}

\subsection{Benchmarks}

Since our goal is to create useful representations of neural images in an unsupervised way, we focused on single- and across-subject action recognition. 
Specifically, we trained our neural decoder $f_{n}$ along with the others without using any action labels. Then, freezing the neural encoder parameters, we trained a linear model on the encoded features,  which is an evaluation protocol widely used in the field \cite{simclr, Lin_2020, He_2020_CVPR, Dave2021TCLRTC}. We used either half or all action labels. We mention the specifics of the train-test split in the Supplementary Material.

\parag{Single-Subject Action Recognition.} For each subject, we trained and tested a simple linear classifier \textit{independently} on the learned representations to predict action labels. We assume that we are given action labels on the subject we are testing.  In \textbf{Table~\ref{tab:accuracy}} we report aggregated results.

\parag{Across-Subject Action Recognition.} We trained linear classifiers on N-1 subjects simultaneously and tested on the left-out one. Therefore, we assume we do not have action labels for the target subject. We repeated the experiment for each individual and report the mean accuracy in \textbf{Table~\ref{tab:accuracy}} and \textbf{Table~\ref{tab:results_h36m}}.

\parag{Identity Recognition.} As a sanity check, we attempted to classify subject identity among the individuals given the learned representations. We again used a linear classifier to test the domain invariance of the learned representations. 
In the case that the learned representations are domain (subject) invariant, we expect that the linear classifier will not be able to detect the domain of the representations, resulting in a lower identity recognition accuracy. Identity recognition results are rerported in \textbf{Table~\ref{tab:accuracy}} and \textbf{Table~\ref{tab:results_h36m}}.
\subsection{Results}
\vspace{10pt}

\parag{Single-Subject Action Recognition on M2CP.}

For the Single-Subject baseline, joint modeling of common latent space out-performed supervised models by a large margin, even when the linear classifier was trained on the action labels of the tested animal. Our swapping and neural augmentations resulted in an accuracy boost when compared with a simple contrastive learning method, SimCLR\cite{simclr}. Although regression-based methods can extract behavioral information from the neural data, they do not produce discriminative features. When combined with the proposed set of augmentations, our method performs better than previous neural decoding models because it extracts richer features thanks to a better unsupervised pretraining step. Domain adaptation techniques do not result in a significant difference in the single-subject baseline; the domain gap in a single animal is smaller than between animals.

\parag{Across-Subject Action Recognition on M2CP.}

We show that supervised models do not generalize across animals, because each nervous system is unique. Before using the proposed augmentations, the contrastive method SimCLR performed worse than convolutional and recurrent regression-based methods including the current state-of-art BehaveNet~\cite{behavenet}. This was due to large domain gap between animals in the latent embeddings \textbf{(Fig.~\ref{fig:domain}C)}. Although the domain adaptation methods MMD (Maximum Mean Discrepancy) and GRL (Gradient Reversal Layer) close the domain gap when used with contrastive learning, they do not position semantically similar points near one another \textbf{(Fig.~\ref{fig:domain}D)}. As a result, domain adaptation-based methods do not result in significant improvements in the across-subject action recognition task. Although regression-based methods suffer less from the domain gap problem, they do not produce representations that are as discriminative as contrastive learning-based methods. Our proposed set of augmentations close the domain gap, while improving the action recognition baseline for self-supervised methods, for both single-subject and across-subject tasks \textbf{(Fig.~\ref{fig:domain}E)}.

\parag{Action Recognition on ECoG Motion vs Rest.} 

As shown at the bottom of \textbf{Table~\ref{tab:results_h36m}}, our approach significantly lowers the identity information in ECoG embeddings, while significantly increasing across-subject action recognition accuracy compared to the regression and multi-modal SimCLR baselines. Low supervised accuracy confirms a strong domain gap across individuals. Note that uni-modal contrastive modeling of ECoG recordings (SimCLR (ECoG)) does not yield strong across-subject action classification accuracy because uni-modal modeling cannot deal with the large domain gap in the learned representations.

\parag{Human Action Recognition on H3.6M.}
We observe in \textbf{Table~\ref{tab:results_h36m}} that, similar to the previous datasets, the low performance of the supervised baseline and the uni-modal modeling of RGB images (SimCLR (RGB)) are due to the domain-gap in the across-subject benchmark. This observation is confirmed by the high identity recognition of these models. %
Our swapping augmentation strongly improves compared to the regression and multi-modal contrastive (SimCLR) baselines. Similar to the previous datasets, uni-modal contrastive training cannot generalize across subjects, due to the large domain gap.

\subsection{Ablation Study}

\begin{table}[!h]
\setlength{\tabcolsep}{0.3em}

\centering
\footnotesize{
    \begin{tabular}{l|l|l|l}
    \toprule
      &   \textbf{Single}  & \textbf{Across}  & \textbf{Identity}\\
        \textbf{Method} &   \textbf{Subj.$\uparrow$}  & \textbf{Subj.$\uparrow$}  & \textbf{Recog. $\downarrow$}\\
        \midrule
         w/ Swapping Augmentation & \textcolor{ForestGreen}{$\blacktriangle$} + 2.7 & \textcolor{ForestGreen}{$\blacktriangle$} + 7.9 & \textcolor{ForestGreen}{$\blacktriangledown$}  \hspace{1pt}-63.0 \\
         + w/ Calcium Augmentation  & \textcolor{ForestGreen}{$\blacktriangle$} + 2.1 &  \textcolor{ForestGreen}{$\blacktriangle$} + 2.7 & $\blacksquare$  +1.2  \\
         + w/ Mix Augmentation & \textcolor{ForestGreen}{$\blacktriangle$} + 1.1 & \textcolor{ForestGreen}{$\blacktriangle$} + 1.2 & $\blacksquare$   \hspace{1pt}-0.8 \\
    \bottomrule
    \end{tabular}
\caption{\textbf{Ablation Study.} Showing the effect of different augmentations on single-subject, across-subject and identity recognition benchmarks.}
\label{tab:ablation}
}
\vspace{-1em}
\end{table}

We compare the individual contributions of different augmentations proposed in our method. We report these results in \textbf{Table~\ref{tab:ablation}}. We observe that all augmentations contribute to single- and across-subject benchmarks. Our swapping augmentation strongly affects the across-subject benchmark, while at the same time greatly decreasing the domain gap, as quantified by the identity recognition result. Other augmentations have minimal effects on the domain gap, as they only slightly affect the identity recognition benchmark.

\section{Conclusion}

We have introduced an unsupervised neural action representation framework for neural imaging and behavioral videography data. We extended previous methods by incorporating a new swapping based domain adaptation technique which we have shown to be useful on three very different multimodal datasets, together with a set of domain-specific neural augmentations.  Two of these datasets are publicly available. We created the third dataset, which we call MC2P, by recording video and neural data for \textit{\drosophila} and will release it publicly to speed-up the development of self-supervised methods in neuroscience. We hope our work will help the development of effective brain machine interface and neural decoding algorithms. In future work, we plan to disentangle remaining long-term non-behavioral information that has a global effect on neural data, such as hunger or thirst, and test our method on different neural recording modalities. As a potential negative impact, we assume that once neural data is taken without consent, our method can be used to extract private information.


\clearpage

{\small
\bibliographystyle{ieee_fullname}
\bibliography{top.bbl}

\begin{thebibliography}{10}\itemsep=-1pt

\bibitem{Multiscale}
Hamidreza Abbaspourazad, Mahdi Choudhury, Yan~T. Wong, Bijan Pesaran, and
  Maryam~M. Shanechi.
\newblock Multiscale low-dimensional motor cortical state dynamics predict
  naturalistic reach-and-grasp behavior.
\newblock {\em Nature Communications}, 12(1):607, 2021.

\bibitem{aymanns21ofco}
Florian Aymanns.
\newblock ofco: optical flow motion correction.
\newblock {\em https://doi.org/10.5281/zenodo.5518800}, Sep 2021.

\bibitem{aymanns21utils2p}
Florian Aymanns.
\newblock utils2p.
\newblock {\em https://doi.org/10.5281/zenodo.5501119}, Sep 2021.

\bibitem{bahdanau}
Dzmitry Bahdanau, {Kyung Hyun} Cho, and Yoshua Bengio.
\newblock Neural machine translation by jointly learning to align and
  translate.
\newblock In {\em Proceedings of the International Conference on Machine
  Learning (ICML)}, 2015.

\bibitem{Bala}
Praneet~C. Bala, Benjamin~R. Eisenreich, Seng Bum~Michael Yoo, Benjamin~Y.
  Hayden, Hyun~Soo Park, and Jan Zimmermann.
\newblock Automated markerless pose estimation in freely moving macaques with
  openmonkeystudio.
\newblock {\em Nature Communications}, 11(1):4560, 2020.

\bibitem{behavenet}
Eleanor Batty, Matthew Whiteway, Shreya Saxena, Dan Biderman, Taiga Abe, Simon
  Musall, Winthrop Gillis, Jeffrey Markowitz, Anne Churchland, John~P
  Cunningham, et~al.
\newblock Behavenet: nonlinear embedding and bayesian neural decoding of
  behavioral videos.
\newblock In {\em Advances in Neural Information Processing Systems (NeurIPS)},
  2019.

\bibitem{Cande}
Jessica Cande, Shigehiro Namiki, Jirui Qiu, Wyatt Korff, Gwyneth~M Card,
  Joshua~W Shaevitz, David~L Stern, and Gordon~J Berman.
\newblock {Optogenetic dissection of descending behavioral control in
  Drosophila}.
\newblock {\em eLife}, 7:970, 2018.

\bibitem{chen18}
K. Chen, P. Gabriel, A. Alasfour, C. Gong, W.~K. Doyle, O. Devinsky, D.
  Friedman, P. Dugan, L. Melloni, T. Thesen, D. Gonda, S. Sattar, S. Wang, and
  V. Gilja.
\newblock {Patient-Specific Pose Estimation in Clinical Environments}.
\newblock {\em IEEE Journal of Translational Engineering in Health and
  Medicine}, 6:1--11, 2018.

\bibitem{chen2019temporal}
Min-Hung Chen, Zsolt Kira, Ghassan AlRegib, Jaekwon Yoo, Ruxin Chen, and Jian
  Zheng.
\newblock Temporal attentive alignment for large-scale video domain adaptation.
\newblock In {\em Proceedings of the IEEE International Conference on Computer
  Vision (ICCV)}, 2019.

\bibitem{simclr}
Ting Chen, Simon Kornblith, Mohammad Norouzi, and Geoffrey Hinton.
\newblock A simple framework for contrastive learning of visual
  representations.
\newblock In {\em Proceedings of the International Conference on Machine
  Learning (ICML)}, 2020.

\bibitem{chen2018imaging}
{Chen C-L, Hermans L}, Meera~C Viswanathan, Denis Fortun, Florian Aymanns,
  Michael Unser, Anthony Cammarato, Michael~H Dickinson, and Pavan Ramdya.
\newblock Imaging neural activity in the ventral nerve cord of behaving adult
  drosophila.
\newblock {\em Nature communications}, 9(1):4390, 2018.

\bibitem{COLLINGER201884}
Jennifer~L. Collinger, Robert~A. Gaunt, and Andrew~B. Schwartz.
\newblock Progress towards restoring upper limb movement and sensation through
  intracortical brain-computer interfaces.
\newblock {\em Current Opinion in Biomedical Engineering}, 8:84--92, 2018.
\newblock Neural Engineering/ Novel Biomedical Technologies: Neuromodulation.

\bibitem{Dave2021TCLRTC}
I. Dave, Rohit Gupta, M.~N. Rizve, and M. Shah.
\newblock {TCLR}: Temporal contrastive learning for video representation.
\newblock {\em arXiv}, 2021.

\bibitem{dombeck}
Daniel~A. Dombeck, Anton~N. Khabbaz, Forrest Collman, Thomas~L. Adelman, and
  David~W. Tank.
\newblock Imaging large-scale neural activity with cellular resolution in
  awake, mobile mice.
\newblock {\em Neuron}, 56(1):43 -- 57, 2007.

\bibitem{Ecker2010c}
A.~S. Ecker, P. Berens, G.~A. Keliris, M. Bethge, N.~K. Logothetis, and A.~S.
  Tolias.
\newblock Decorrelated neuronal firing in cortical microcircuits.
\newblock {\em Science}, 327(5965):584--587, 2010.

\bibitem{stablebrainmachine}
Ali Farshchian, Juan Gallego, Joseph Cohen, Y. Bengio, Lee Miller, and Sara
  Solla.
\newblock Adversarial domain adaptation for stable brain-machine interfaces.
\newblock {\em arXiv}, 2018.

\bibitem{GRL}
Yaroslav Ganin and Victor Lempitsky.
\newblock Unsupervised domain adaptation by backpropagation.
\newblock In {\em Proceedings of the International Conference on Machine
  Learning (ICML)}, 2015.

\bibitem{GANZER2020763}
Patrick~D. Ganzer, Samuel~C. Colachis, Michael~A. Schwemmer, David~A.
  Friedenberg, Collin~F. Dunlap, Carly~E. Swiftney, Adam~F. Jacobowitz, Doug~J.
  Weber, Marcia~A. Bockbrader, and Gaurav Sharma.
\newblock Restoring the sense of touch using a sensorimotor demultiplexing
  neural interface.
\newblock {\em Cell}, 181(4):763--773.e12, 2020.

\bibitem{Gao2016LinearDN}
Yuanjun Gao, Evan Archer, L. Paninski, and J. Cunningham.
\newblock Linear dynamical neural population models through nonlinear
  embeddings.
\newblock In {\em Advances in Neural Information Processing Systems (NeurIPS)},
  2016.

\bibitem{MLfordecoding}
Joshua~I. Glaser, Ari~S. Benjamin, Raeed~H. Chowdhury, Matthew~G. Perich,
  Lee~E. Miller, and Konrad~P. Kording.
\newblock Machine learning for neural decoding.
\newblock {\em eNeuro}, 7(4), 2020.

\bibitem{lp3d}
Adam Gosztolai, Semih G{\"u}nel, Victor Lobato-R{\'\i}os, Marco Pietro~Abrate,
  Daniel Morales, Helge Rhodin, Pascal Fua, and Pavan Ramdya.
\newblock Liftpose3d, a deep learning-based approach for transforming
  two-dimensional to three-dimensional poses in laboratory animals.
\newblock {\em Nature Methods}, 18(8):975--981, 2021.

\bibitem{couzin}
Jacob~M Graving, Daniel Chae, Hemal Naik, Liang Li, Benjamin Koger, Blair~R
  Costelloe, and Iain~D Couzin.
\newblock Deepposekit, a software toolkit for fast and robust animal pose
  estimation using deep learning.
\newblock {\em eLife}, 8:e47994, 2019.

\bibitem{MMD}
Arthur Gretton, Karsten~M. Borgwardt, Malte~J. Rasch, Bernhard Sch{\"{o}}lkopf,
  and Alexander~J. Smola.
\newblock A kernel method for the two-sample-problem.
\newblock In {\em Proceedings of Advances in Neural Information Processing
  Systems (NeurIPS)}, 2006.

\bibitem{Gunel19a}
S. G{\"u}nel, H. Rhodin, D. Morales, J. Campagnolo, P. Ramdya, and P. Fua.
\newblock {Deepfly3D, a Deep Learning-Based Approach for 3D Limb and Appendage
  Tracking in Tethered, Adult Drosophila}.
\newblock {\em eLife}, 8:e48571, 2019.

\bibitem{He_2020_CVPR}
Kaiming He, Haoqi Fan, Yuxin Wu, Saining Xie, and Ross Girshick.
\newblock Momentum contrast for unsupervised visual representation learning.
\newblock In {\em Proceedings of the IEEE/CVF Conference on Computer Vision and
  Pattern Recognition (CVPR)}, June 2020.

\bibitem{batchnorm}
Sergey Ioffe and Christian Szegedy.
\newblock Batch normalization: Accelerating deep network training by reducing
  internal covariate shift.
\newblock In {\em Proceedings of the International Conference on Machine
  Learning (ICML)}, 2015.

\bibitem{h36m_pami}
Catalin Ionescu, Dragos Papava, Vlad Olaru, and Cristian Sminchisescu.
\newblock Human3.6m: Large scale datasets and predictive methods for 3d human
  sensing in natural environments.
\newblock {\em IEEE Transactions on Pattern Analysis and Machine Intelligence},
  36(7):1325--1339, 2014.

\bibitem{Ionescu14a}
C. Ionescu, I. Papava, V. Olaru, and C. Sminchisescu.
\newblock {{Human3.6M}: Large Scale Datasets and Predictive Methods for 3D
  Human Sensing in Natural Environments}.
\newblock {\em IEEE Transactions on Pattern Analysis and Machine Intelligence},
  2014.

\bibitem{robertZebraFish20}
Robert~Evan Johnson, Scott Linderman, Thomas Panier, Caroline~Lei Wee, Erin
  Song, Kristian~Joseph Herrera, Andrew Miller, and Florian Engert.
\newblock Probabilistic models of larval zebrafish behavior reveal structure on
  many scales.
\newblock {\em Current Biology}, 30(1):70--82.e4, 2020.

\bibitem{kangcontrastive}
Guoliang Kang, Lu Jiang, Yunchao Wei, Yi Yang, and Alexander~G Hauptmann.
\newblock Contrastive adaptation network for single-and multi-source domain
  adaptation.
\newblock {\em IEEE Transactions on Pattern Analysis and Machine Intelligence
  (TPAMI)}, 2020.

\bibitem{anipose}
Pierre Karashchuk, Katie~L. Rupp, Evyn~S. Dickinson, Sarah Walling-Bell,
  Elischa Sanders, Eiman Azim, Bingni~W. Brunton, and John~C. Tuthill.
\newblock Anipose: A toolkit for robust markerless 3d pose estimation.
\newblock {\em Cell}, 36(13):109730, 2021.

\bibitem{adam}
Diederik~P. Kingma and Jimmy Ba.
\newblock Adam: A method for stochastic optimization.
\newblock In {\em International Conference on Learning Representations,
  (ICLR)}, 2015.

\bibitem{eegselfsupervised}
Demetres Kostas, Stéphane Aroca-Ouellette, and Frank Rudzicz.
\newblock Bendr: Using transformers and a contrastive self-supervised learning
  task to learn from massive amounts of eeg data.
\newblock {\em Frontiers in Human Neuroscience}, 15:253, 2021.

\bibitem{decodingannotation}
Karine Lacourse, Ben Yetton, Sara Mednick, and Simon~C. Warby.
\newblock Massive online data annotation, crowdsourcing to generate high
  quality sleep spindle annotations from eeg data.
\newblock {\em Scientific Data}, 7(1):190, 2020.

\bibitem{closed-loop}
Celia K~S Lau, Meghan Jelen, and Michael~D Gordon.
\newblock A closed-loop optogenetic screen for neurons controlling feeding in
  drosophila.
\newblock {\em G3 (Bethesda)}, 11(5), 05 2021.

\bibitem{deepinterpolation}
J{\'e}r{\^o}me Lecoq, Michael Oliver, Joshua~H. Siegle, Natalia Orlova, and
  Christof Koch.
\newblock Removing independent noise in systems neuroscience data using
  deepinterpolation.
\newblock {\em bioRxiv}, 2020.

\bibitem{li2020deformation}
Siyuan Li, Semih G\"unel, Mirela Ostrek, Pavan Ramdya, Pascal Fua, and Helge
  Rhodin.
\newblock Deformation-aware unpaired image translation for pose estimation on
  laboratory animals.
\newblock In {\em Proceedings of the IEEE/CVF Conference on Computer Vision and
  Pattern Recognition (CVPR)}, 2020.

\bibitem{domainadaptneuraldecoding}
Wei Li, Shaohua Ji, Xi Chen, Bo Kuai, Jiping He, Zhang Peng, and Qiang Li.
\newblock Multi-source domain adaptation for decoder calibration of
  intracortical brain-machine interface.
\newblock {\em Journal of neural engineering}, 17, 10 2020.

\bibitem{Lin_2020}
Lilang Lin, Sijie Song, Wenhan Yang, and Jiaying Liu.
\newblock {MS2L}: Multi-task self-supervised learning for skeleton based action
  recognition.
\newblock In {\em Proceedings of the ACM International Conference on
  Multimedia}, 2020.

\bibitem{pmlr-v54-linderman17a}
Scott Linderman, Matthew Johnson, Andrew Miller, Ryan Adams, David Blei, and
  Liam Paninski.
\newblock {Bayesian Learning and Inference in Recurrent Switching Linear
  Dynamical Systems}.
\newblock In {\em Proceedings of the International Conference on Artificial
  Intelligence and Statistics (AISTATS)}, 2017.

\bibitem{Linderman621540}
Scott Linderman, Annika Nichols, David Blei, Manuel Zimmer, and Liam Paninski.
\newblock Hierarchical recurrent state space models reveal discrete and
  continuous dynamics of neural activity in c. elegans.
\newblock {\em bioRxiv}, 2019.

\bibitem{liu2020snce}
Yuejiang Liu, Qi Yan, and Alexandre Alahi.
\newblock Social nce: Contrastive learning of socially-aware motion
  representations.
\newblock {\em arXiv}, 2020.

\bibitem{eegcontrastive}
Mostafa~Neo Mohsenvand, Mohammad~Rasool Izadi, and Pattie Maes.
\newblock Contrastive representation learning for electroencephalogram
  classification.
\newblock In {\em Proceedings of the Machine Learning for Health NeurIPS
  Workshop}, 2020.

\bibitem{munro20multi}
Jonathan Munro and Dima Damen.
\newblock {M}ulti-modal {D}omain {A}daptation for {F}ine-grained {A}ction
  {R}ecognition.
\newblock In {\em Proceedings of the IEEE/CVF Conference on Computer Vision and
  Pattern Recognition (CVPR)}, 2020.

\bibitem{musall19}
Simon Musall, Matthew~T. Kaufman, Ashley~L. Juavinett, Steven Gluf, and Anne~K.
  Churchland.
\newblock Single-trial neural dynamics are dominated by richly varied
  movements.
\newblock {\em Nature Neuroscience}, 22(10):1677--1686, 2019.

\bibitem{supervised1}
Sho Nakagome, Trieu~Phat Luu, Yongtian He, Akshay~Sujatha Ravindran, and
  Jose~L. Contreras-Vidal.
\newblock An empirical comparison of neural networks and machine learning
  algorithms for eeg gait decoding.
\newblock {\em Nature Scientific Reports}, 10(1):4372, 2020.

\bibitem{Nassar2019TreeStructuredRS}
Josue Nassar, Scott~W. Linderman, M. Bugallo, and Il-Su Park.
\newblock Tree-structured recurrent switching linear dynamical systems for
  multi-scale modeling.
\newblock {\em arXiv}, 2019.

\bibitem{dlc}
Tanmay Nath, Alexander Mathis, An~Chi Chen, Amir Patel, Matthias Bethge, and
  Mackenzie~Weygandt Mathis.
\newblock Using deeplabcut for 3d markerless pose estimation across species and
  behaviors.
\newblock {\em Nature Protocols}, 14(7):2152--2176, 2019.

\bibitem{berman21}
Katherine Overman, Daniel Choi, Kawai Leung, Joshua Shaevitz, and Gordon
  Berman.
\newblock Measuring the repertoire of age-related behavioral changes in
  drosophila melanogaster.
\newblock {\em bioRxiv}, 2021.

\bibitem{pan2021videomoco}
Tian Pan, Yibing Song, Tianyu Yang, Wenhao Jiang, and Wei Liu.
\newblock Videomoco: Contrastive video representation learning with temporally
  adversarial examples.
\newblock In {\em Proceedings of the IEEE/CVF Conference on Computer Vision and
  Pattern Recognition (CVPR)}, 2021.

\bibitem{lfads}
Chethan Pandarinath, Daniel~J. O'Shea, Jasmine Collins, Rafal Jozefowicz,
  Sergey~D. Stavisky, Jonathan~C. Kao, Eric~M. Trautmann, Matthew~T. Kaufman,
  Stephen~I. Ryu, Leigh~R. Hochberg, Jaimie~M. Henderson, Krishna~V. Shenoy,
  L.~F. Abbott, and David Sussillo.
\newblock Inferring single-trial neural population dynamics using sequential
  auto-encoders.
\newblock {\em Nature Methods}, 15(10):805--815, 2018.

\bibitem{zebradataset}
Malte Pedersen, Joakim~Bruslund Haurum, Stefan~Hein Bengtson, and Thomas~B.
  Moeslund.
\newblock 3d-zef: A 3d zebrafish tracking benchmark dataset.
\newblock In {\em IEEE/CVF Conference on Computer Vision and Pattern
  Recognition (CVPR)}, June 2020.

\bibitem{pei2021neural}
Felix Pei, Joel Ye, David Zoltowski, Anqi Wu, Raeed~H. Chowdhury, Hansem Sohn,
  Joseph~E. O'Doherty, Krishna~V. Shenoy, Matthew~T. Kaufman, Mark Churchland,
  Mehrdad Jazayeri, Lee~E. Miller, Jonathan Pillow, Il~Memming Park, Eva~L.
  Dyer, and Chethan Pandarinath.
\newblock Neural latents benchmark '21: Evaluating latent variable models of
  neural population activity, 2021.

\bibitem{quantify}
Talmo~D. Pereira, Joshua~W. Shaevitz, and Mala Murthy.
\newblock Quantifying behavior to understand the brain.
\newblock {\em Nature Neuroscience}, 23(12):1537--1549, 2020.

\bibitem{sleap}
Talmo~D. Pereira, Nathaniel Tabris, Junyu Li, Shruthi Ravindranath, Eleni~S.
  Papadoyannis, Z.~Yan Wang, David~M. Turner, Grace McKenzie-Smith, Sarah~D.
  Kocher, Annegret~L. Falkner, Joshua~W. Shaevitz, and Mala Murthy.
\newblock Sleap: Multi-animal pose tracking.
\newblock {\em bioRxiv}, 2020.

\bibitem{peterson_2021}
Steven Peterson.
\newblock Ecog and arm position during movement and rest, Sep 2021.

\bibitem{labelingneural}
Steven~M. Peterson, Rajesh P.~N. Rao, and Bingni~W. Brunton.
\newblock Learning neural decoders without labels using multiple data streams.
\newblock {\em bioRxiv}, 2021.

\bibitem{pnevmatikakis2013bayesian}
Eftychios~A. Pnevmatikakis, Josh Merel, Ari Pakman, and Liam Paninski.
\newblock Bayesian spike inference from calcium imaging data.
\newblock {\em arXiv}, 2013.

\bibitem{subtrate}
Alice~A. Robie, Jonathan Hirokawa, Austin~W. Edwards, Lowell~A. Umayam, Allen
  Lee, Mary~L. Phillips, Gwyneth~M. Card, Wyatt Korff, Gerald~M. Rubin,
  Julie~H. Simpson, Michael~B. Reiser, and Kristin Branson.
\newblock Mapping the neural substrates of behavior.
\newblock {\em Cell}, 170(2):393--406.e28, 2017.

\bibitem{Rupprecht21}
Peter Rupprecht, Stefano Carta, Adrian Hoffmann, Mayumi Echizen, Antonin Blot,
  Alex~C. Kwan, Yang Dan, Sonja~B. Hofer, Kazuo Kitamura, Fritjof Helmchen, and
  Rainer~W. Friedrich.
\newblock A database and deep learning toolbox for noise-optimized, generalized
  spike inference from calcium imaging.
\newblock {\em Nature Neuroscience}, 24(9):1324--1337, 2021.

\bibitem{subspace}
Omid~G. Sani, Hamidreza Abbaspourazad, Yan~T. Wong, Bijan Pesaran, and
  Maryam~M. Shanechi.
\newblock Modeling behaviorally relevant neural dynamics enabled by
  preferential subspace identification.
\newblock {\em Nature Neuroscience}, 24(1):140--149, 2021.

\bibitem{biologicalinsight}
Omid~G Sani, Yuxiao Yang, Morgan~B Lee, Heather~E Dawes, Edward~F Chang, and
  Maryam~M Shanechi.
\newblock Mood variations decoded from multi-site intracranial human brain
  activity.
\newblock {\em Nature Biotechnology}, 36(10):954--961, 2018.

\bibitem{Seelig}
Johannes~D Seelig, M~Eugenia Chiappe, Gus~K Lott, Anirban Dutta, Jason~E
  Osborne, Michael~B Reiser, and Vivek Jayaraman.
\newblock {Two-photon calcium imaging from head-fixed Drosophila during
  optomotor walking behavior}.
\newblock {\em Nature Methods}, 7(7):535--540, 2010.

\bibitem{Segalin2020}
Cristina Segalin, Jalani Williams, Tomomi Karigo, May Hui, Moriel Zelikowsky,
  Jennifer~J. Sun, Pietro Perona, David~J. Anderson, and Ann Kennedy.
\newblock The mouse action recognition system (mars): a software pipeline for
  automated analysis of social behaviors in mice.
\newblock {\em bioRxiv}, 2020.

\bibitem{neuralrecordingtech}
Krishna~V. Shenoy and Jonathan~C. Kao.
\newblock Measurement, manipulation and modeling of brain-wide neural
  population dynamics.
\newblock {\em Nature Communications}, 12(1):633, 2021.

\bibitem{SINGH2021109199}
Satpreet~H. Singh, Steven~M. Peterson, Rajesh~P.N. Rao, and Bingni~W. Brunton.
\newblock Mining naturalistic human behaviors in long-term video and neural
  recordings.
\newblock {\em Journal of Neuroscience Methods}, 358:109199, 2021.

\bibitem{stringer19}
Carsen Stringer, Marius Pachitariu, Nicholas Steinmetz, Charu~Bai Reddy, Matteo
  Carandini, and Kenneth~D Harris.
\newblock Spontaneous behaviors drive multidimensional, brainwide activity.
\newblock {\em Science}, 364(6437):255--255, 2019.

\bibitem{su2020predict}
Kun Su, Xiulong Liu, and Eli Shlizerman.
\newblock Predict \& cluster: Unsupervised skeleton based action recognition.
\newblock In {\em Proceedings of the IEEE/CVF Conference on Computer Vision and
  Pattern Recognition (CVPR)}, 2020.

\bibitem{task_programming}
Jennifer~J Sun, Ann Kennedy, Eric Zhan, David~J Anderson, Yisong Yue, and
  Pietro Perona.
\newblock Task programming: Learning data efficient behavior representations.
\newblock In {\em Proceedings of the IEEE/CVF Conference on Computer Vision and
  Pattern Recognition (CVPR)}, 2021.

\bibitem{wirelesshuman}
Uros Topalovic, Zahra~M. Aghajan, Diane Villaroman, Sonja Hiller, Leonardo
  Christov-Moore, Tyler~J. Wishard, Matthias Stangl, Nicholas~R. Hasulak,
  Cory~S. Inman, Tony~A. Fields, Vikram~R. Rao, Dawn Eliashiv, Itzhak Fried,
  and Nanthia Suthana.
\newblock Wireless programmable recording and stimulation of deep brain
  activity in freely moving humans.
\newblock {\em Neuron}, 108(2):322--334.e9, 2020.

\bibitem{largescale}
Anne~E. Urai, Brent Doiron, Andrew~M. Leifer, and Anne~K. Churchland.
\newblock Large-scale neural recordings call for new insights to link brain and
  behavior.
\newblock {\em arXiv}, 2021.

\bibitem{neuromusculardisease}
Kota Utsumi, Kouji Takano, Yoji Okahara, Tetsuo Komori, Osamu Onodera, and
  Kenji Kansaku.
\newblock Operation of a p300-based brain-computer interface in patients with
  duchenne muscular dystrophy.
\newblock {\em Scientific Reports}, 8(1):1753, 2018.

\bibitem{oord2019representation}
Aaron van~den Oord, Yazhe Li, and Oriol Vinyals.
\newblock Representation learning with contrastive predictive coding.
\newblock {\em arXiv}, 2019.

\bibitem{Wang2018AJILEMP}
X. Wang, Ali Farhadi, Rajesh P.~N. Rao, and Bingni~W. Brunton.
\newblock Ajile movement prediction: Multimodal deep learning for natural human
  neural recordings and video.
\newblock In {\em Association for the Advancement of Artificial Intelligence
  (AAAI)}, 2018.

\bibitem{wei-zou-2019-eda}
Jason Wei and Kai Zou.
\newblock {EDA}: Easy data augmentation techniques for boosting performance on
  text classification tasks.
\newblock In {\em Proceedings of the Conference on Empirical Methods in Natural
  Language Processing and the International Joint Conference on Natural
  Language Processing (EMNLP-IJCNLP)}, 2019.

\bibitem{bmi}
Shixian Wen, Allen Yin, Po-He Tseng, Laurent Itti, Mikhail~A. Lebedev, and
  Miguel Nicolelis.
\newblock Capturing spike train temporal pattern with wavelet average
  coefficient for brain machine interface.
\newblock {\em Scientific Reports}, 11(1):19020, 2021.

\bibitem{graphpose}
Anqi Wu, Estefany~Kelly Buchanan, Matthew Whiteway, Michael Schartner, Guido
  Meijer, Jean-Paul Noel, Erica Rodriguez, Claire Everett, Amy Norovich, Evan
  Schaffer, Neeli Mishra, C.~Daniel Salzman, Dora Angelaki, Andr\'{e}s
  Bendesky, The International Brain~Laboratory The International
  Brain~Laboratory, John~P Cunningham, and Liam Paninski.
\newblock Deep graph pose: a semi-supervised deep graphical model for improved
  animal pose tracking.
\newblock In {\em Advances in Neural Information Processing Systems (NeurIPS)},
  2020.

\bibitem{xu2021aligning}
Yuecong Xu, Jianfei Yang, Haozhi Cao, Kezhi Mao, Jianxiong Yin, and Simon See.
\newblock Aligning correlation information for domain adaptation in action
  recognition.
\newblock {\em arXiv}, 2021.

\bibitem{Yuan_2021_CVPR}
Xin Yuan, Zhe Lin, Jason Kuen, Jianming Zhang, Yilin Wang, Michael Maire,
  Ajinkya Kale, and Baldo Faieta.
\newblock Multimodal contrastive training for visual representation learning.
\newblock In {\em Proceedings of the IEEE/CVF Conference on Computer Vision and
  Pattern Recognition (CVPR)}, 2021.

\bibitem{cutmix}
Sangdoo Yun, Dongyoon Han, Seong~Joon Oh, Sanghyuk Chun, Junsuk Choe, and
  Youngjoon Yoo.
\newblock Cutmix: Regularization strategy to train strong classifiers with
  localizable features.
\newblock In {\em International Conference on Computer Vision (ICCV)}, 2019.

\bibitem{mixup}
Hongyi Zhang, Moustapha Cisse, Yann~N. Dauphin, and David Lopez-Paz.
\newblock mixup: Beyond empirical risk minimization.
\newblock In {\em International Conference on Learning Representations (ICLR)},
  2018.

\bibitem{zhang2020contrastive}
Yuhao Zhang, Hang Jiang, Yasuhide Miura, Christopher~D. Manning, and Curtis~P.
  Langlotz.
\newblock Contrastive learning of medical visual representations from paired
  images and text.
\newblock {\em arXiv}, 2020.

\end{thebibliography}
}

\clearpage

\appendix

\begin{strip}%
 \centering
 \LARGE Appendix for Overcoming the Domain Gap in \\ Neural Action Representations \\[1.5em]
 \normalsize
\end{strip}

\section{Human Actions}

We apply multi-modal contrastive learning on windows of time series and RGB videos. We make the analogy that, similar to the neural data, RGB videos from different view angles show a domain gap although they are tied to the same 3D pose. Therefore, to test our method, we select three individuals with different camera angles where all actors perform the same three actions. We test domain adaptation using the Across-Subject benchmark, where we train our linear action classifier on labels of one individual and test it on the others. We repeat the same experiment three times and report the mean results. We show the results of Across-Subject and Identity Recognition in \textbf{Table~\ref{tab:results_h36m}}. 

For preprocessing, we remove global translation and rotation from 3D poses by subtracting the root joint and then rotating the skeletons to point in the same direction. We use resnet18 for the RGB encoder and a 4 layer convolutional network for the 3D pose encoder. We use S1, S5 and S7 and all their behaviors for training, except for the three behaviors which we used for testing. For each number, we report three-fold cross-validation results.

\begin{figure*}[t]%
  \centering
  \includegraphics[width=1\textwidth]%
  {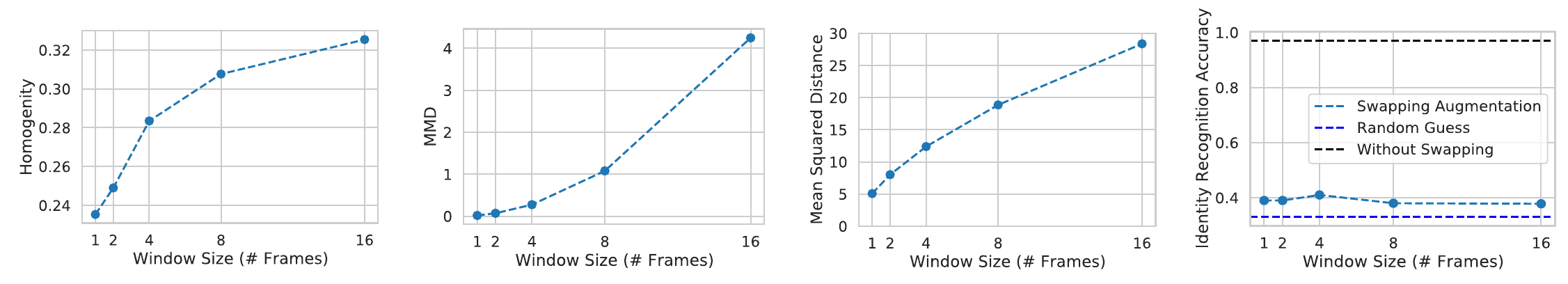}
  \caption{\textbf{Changing window size for the swapping of behavioral modality on the MC2P dataset.} Statistics of the behavioral modality as a function of changing the window size. Decreasing the window size increases clustering homogeneity and Mean Maximum Discrepancy (MMD) when applied to the raw data, therefore suggesting higher quality swapping in individual poses instead of sequences of poses. Swapping augmentation with a smaller window size lowers the degree of perturbation, quantified by Mean Squared Distance. However, identity recognition accuracy does not change considerably when swapping is done with different window sizes. }
  \label{fig:swapping_statistics}
\end{figure*}

\section{Dataset Details} 

\paragraph{Dataset Collection.} Here we provide a more detailed technical explanation of the experimental dataset. Transgenic female \textit{Drosophila melanogaster} flies aged 2-4 days post-eclosion were selected for experiments. They were raised on a 12h:12h day, night light cycle and recorded in either the morning or late afternoon Zeitgeber time. Flies expressed both GCaMP6s and tdTomato in all brain neurons as delineated by otd-Gal4 expression,

{ \hspace{-15pt} \footnotesize
($;\frac{Otd-nls:FLPo (attP40)}{P{20XUAS-IVS-GCaMP6s}attP40};\frac{R57C10-GAL4, tub>GAL80>}{P{w[+mC]=UAS-tdTom.S}3}.$)
}
The fluorescence of GCaMP6s proteins within the neuron increases when it binds to calcium. There is an increase in intracellular calcium when neurons become active and fire action potentials.
Due to the relatively slow release (as opposed to binding) of calcium by  GCaMP6s molecules, the signal decays exponentially.
We also expressed the red fluorescent protein, tdTomato, in the same neurons as an anatomical fiduciary to be used for neural data registration. This compensates for image deformations and translations during animal movements.
We recorded neural data using a two-photon microscope (ThorLabs, Germany; Bergamo2) by scanning the cervical connective. This neural tissue serves as a conduit between the brain and ventral nerve cord (VNC) \cite{chen2018imaging}.
The brain-only GCaMP6s expression pattern in combination with restrictions of recording to the cervical connective allowed us to record a large population of descending neuron axons while also being certain that none of the axons arose from ascending neurons in the VNC. Because descending neurons are expected to drive ongoing actions \cite{Cande}, this imaging approach has the added benefit of ensuring that the imaged cells should, in principle, relate to paired behavioral data.

\vspace{-10pt}
\paragraph{Neural Pre-Processing.} For neural data processing, data were synchronized using a custom Python package \cite{aymanns21utils2p}.
We then estimated the motion of the neurons using images acquired on the red (tdTomato) PMT channel.
The first image of the first trial was selected as a reference frame to which all other frames were registered.
For image registration, we estimated the vector field describing the motion between two frames. To do this, we numerically solved the optimization problem in \textbf{Eq.~\ref{eq:opticalflow}}, where $w$ is the motion field, $\mathcal{I}_t$ is the image being transformed, $\mathcal{I}_r$ is the reference image, and $\Omega$ is the set of all pixel coordinates \cite{chen2018imaging, aymanns21ofco}.
\begin{align}
    \label{eq:opticalflow}
    \hat{w} = argmin_w \sum_{x\in\Omega} ||\mathcal{I}_t(x + w(x)) &- \mathcal{I}_r(x)||^{2}_{2} \\ &- \lambda \sum_{x\in\Omega} || \nabla w(x) ||^2_2 \nonumber
\end{align}
A smoothness promoting parameter $\lambda$ was empirically set to 800.
We then applied $\hat{w}$ to the green PMT channel (GCaMP6s).
To denoise the motion corrected green signal, we trained a DeepInterpolation network \cite{deepinterpolation} for nine epochs for each animal and applied it to the rest of the frames.
We only used the first 100 frames of each trial and used the first and last trials as validation data.
The batch size was set to 20 and we used 30 frames before and after the current frame as input.
In order to have a direct correlation between pixel intensity and neuronal activity we applied the following transformation to all neural images $\frac{F - F_0}{F_0} \times 100$, where $F_0$ is the baseline fluorescence in the absence of neural activity. To estimate $F_0$, we used the pixel-wise minimum of a moving average of 15 frames.

\paragraph{Neural Fluorescence Signal Decay.}
The formal relationship between the neural image $\mathbf{n}_t$ and neural activity (underlying neural firings) $\mathbf{s}_t$ can be modeled as a first-order autoregressive process
$$\mathbf{n}_t=\gamma \mathbf{n}_{t-1}+ \alpha \mathbf{s}_t,$$
where $\mathbf{s}_t$ is a binary variable indicating an event at time $t$ (e.g. the neuron firing an action potential).
The amplitudes $\gamma$ and $\alpha$ determine the rate at which the signal decays and the initial response to an event, respectively. In general, $0 < \gamma < 1 $, therefore resulting in an exponential decay of information pertaining to $\mathbf{s}_t$ to be inside of $\mathbf{n}_t$. A single neural image $\mathbf{n}_t$ includes decaying information from previous neural activity, and hence carries information from previous behaviors. For more detailed information on calcium dynamics, see \cite{pnevmatikakis2013bayesian, Rupprecht21}. Assuming no neural firings, $\mathbf{s}_{t}=0$, $\mathbf{n}_{t}$ is given by $\mathbf{n}_{t} = \gamma^{t} \mathbf{n}_{0}$. Therefore, we define the calcium kernel $\mathcal{K}$ as $\mathcal{K}_t = \gamma ^ {t}$.

\paragraph{Dataset Analysis.} We show the distribution of annotations across 7 animals and action duration distribution in \textbf{Supplementary Fig.~\ref{fig:dataset}}. Unlike scripted actions in human datasets, the animal behavior is spontaneous, therefore does not follow a uniform distribution. The average duration of behaviors can also change across behaviors. Walking is the most common behavior and lasts longer than other behaviors.
We visualize the correlation between the neural and behavioral energy in \textbf{Supplementary Fig.~\ref{fig:energy}}.  We quantify the energy as the Euclidean distance between consecutive, vectorized 3D poses. Similarly, for the neural energy, we calculate the Euclidean distance between consecutive images. To be able to compare corresponding energies, we first synchronize neural and behavioral modalities. We then smooth the corresponding time series using Gaussian convolution with kernel size of 11 frames. We observe that there is a strong correlation between the modalities, suggesting large mutual information.

\section{Method Details}
\paragraph{Augmentations.}
Aside from the augmentations mentioned before, for the neural image transformation family $\mathcal{T}_n$, we used a sequential application of Poisson noise and Gaussian blur and color jittering. In contrast with recent work on contrastive visual representation learning, we only applied brightness and contrast adjustments in color jittering because neural images have a single channel that measures calcium indicator fluorescence intensity. We did not apply any cropping augmentation, such as cutout, because action representation is often highly localized and non-redundant (e.g., grooming is associated with the activity of a small set of neurons and thus with only a small number of pixels). We applied the same augmentations to each frame in single sample of neural data. 

For the behavior transformation family $\mathcal{T}_b$, we used a sequential application of scaling, shear, and random temporal and spatial dropping. We did not apply rotation and translation augmentations because the animals were tethered (i.e., restrained from moving freely), and their direction and absolute location were fixed throughout the experiment. We did not use time warping because neural and behavioral information are temporally linked (e.g., fast walking has different neural representations than slow walking).

\paragraph{Swapping Parameters.} We analyze the effects of swapping individual poses, instead of whole motion sequences, through our swapping augmentation in \textbf{Fig.~\ref{fig:swapping_statistics}}. We compare the distribution similarity across individuals when tested on single poses and windows of poses. We observe that the distribution similarity across individuals in behavioral modality is much larger in pose level when compared to the whole motion sequence, therefore making it easier to swap behavioral data in pose level. We quantify the distribution similarity using MMD (Mean Maximum Discrepancy) and Homogeneity metrics. Similarly, swapping individual poses decreases the overall change in the motion sequence, as quantified by the Mean Squared Distance. Yet, the degree to which identity information is hid does not strongly correlate with the window size of swapping. Therefore, overall, suggesting swapping in pose level is better than swapping whole motion sequences. 

\paragraph{Implementation Details:}
For all methods, we initialized the weights of the networks randomly unless otherwise specified. To keep the experiments consistent, we always paired $32$ frames of neural data with $8$ frames of behavioral data. For the neural data, we used a larger time window because the timescale during which dynamic changes occur are smaller. For the paired modalities, we considered data synchronized if their center frames had the same timestamp. We trained contrastive methods for $200$ epochs and set the temperature value $\tau$ to $0.1$. We set the output dimension of $\mathbf{z}_b$ and $\mathbf{z}_n$ to $128$. We used a cosine training schedule with three epochs of warm-up. For non-contrastive methods, we trained for $200$ epochs with a learning rate of $1e-4$, and a weight decay of $1e-5$, using the Adam optimizer \cite{adam}. We ran all experiments using an Intel Core i9-7900X CPU, 32 GB of DDR4 RAM, and a GeForce GTX 1080. Training for a single SimCLR network for 200 epochs took 12 hours. To create train and test splits, we removed two trials from each animal and used them only for testing. We used the architecture shown in \textbf{Supplementary Table {\color{pearDark} 1}} for the neural image and behavioral pose encoder. Each layer except the final fully-connected layer was followed by Batch Normalization and a ReLU activation function \cite{batchnorm}. For the self-attention mechanism in the behavioral encoder \textbf{(Supplementary Table~{\color{pearDark} 1})}, we implement Bahdanau attention~\cite{bahdanau}. Given the set of intermediate behavioral representations $S \in \mathbb{R} ^{T \times D}$, we first calculated,
$$
\mathbf{r}=W_{2} \tanh \left(W_{1} S^{\top}\right) \quad \text { and } \quad \mathbf{a}_{i}=-\log \left(\frac{\exp \left(\mathbf{r}_{i}\right)}{\sum_{j} \exp \left(\mathbf{r}_{j}\right)}\right)
$$
where $W_{1}$ and $W_{2}$ are a set of matrices of shape $\mathbb{R}^{12\times D}$ and $\mathbb{R}^{1\times12}$ respectively. $\mathbf{a}_i$ is the assigned score i-th pose in the sequence of motion. Then the final representation is given by $\sum_{i}^{T} \mathbf{a}_i S_{i}$.

\section{Baseline Methods}

\parag{Supervised:} A feedforward network trained with manually annotated action labels using cross-entropy loss, having neural data as input. We discarded datapoints that did not have associated behavioral labels. For the MLP baseline, we trained a simple three layer MLP with a hidden layer size of 128 neurons with ReLU activation and without batch normalization. 

\parag{Regression (Convolutional):} A fully-convolutional feedforward network trained with MSE loss for behavioral reconstruction task, given the set of neural images. To keep the architectures consistent with the other methods, the average pooling is followed by a projection layer, which is used as the final representation of this model.

\parag{Regression (Recurrent):} This is similar to the one above but the last projection network was replaced with a two-layer GRU module. The GRU module takes as an input the fixed representation of neural images. At each time step, the GRU module predicts a single 3D pose with a total of eight steps to predict the eight poses associated with an input neural image. This model is trained with an MSE loss. We take the input of the GRU module as the final representation of neural encoder. 

\begin{table}[t]
\setlength{\tabcolsep}{2pt}

\caption*{{\bf (a)} First part of the Neural Encoder $f_n$}
\vspace{-10pt}
\scriptsize
\begin{center}
\begin{tabular}[t]{  l r c c r }
 \toprule
 Layer & \# filters & K & S   & Output  \\  
 \midrule
 input  & 1 &   - &  -             & $T\times 128 \times 128 $  \\ 
 conv1  & 2 & (3,3)   & (1,1)     & $T\times 128 \times 128 $  \\ 
 mp2   & -   & (2,2)   & (2,2)     & $T\times 64 \times 64 $  \\ 
 conv3 & 4 & (3,3)   & (1,1)     & $T\times 64 \times 64 $  \\ 
 mp4   & -   & (2,2)   & (2,2)     & $T\times 32 \times 32 $  \\ 
 conv5 & 8 & (3,3)   & (1,1)     & $T\times 32 \times 32 $  \\ 
 mp6   & -   & (2,2)   & (2,2)     & $T\times 16 \times 16 $  \\
 conv7 & 16 & (3,3)   & (1,1)     & $T\times 16 \times 16 $  \\
 mp8   & -   & (2,2)   & (2,2)     & $T\times 8 \times 8 $  \\
 conv9 & 32 & (3,3)   & (1,1)     & $T\times 8 \times 8 $  \\
 mp10   & -   & (2,2)   & (2,2)     & $T\times 4 \times 4 $  \\
 conv11 & 64 & (3,3)   & (1,1)     & $T\times 4 \times 4 $  \\
 mp12   &  -   & (2,2)   & (2,2)     & $T\times 2 \times 2 $  \\
 fc13   & 128  & (1,1)   & (1,1)     & $T\times 1 \times 1 $  \\
 fc14   & 128  & (1,1)   & (1,1) & $T\times 1 \times  1$  \\
 \bottomrule
\end{tabular}

\vspace{10pt}
\caption*{{\bf (a)} Second part of the Neural Encoder $f_n$}
\scriptsize
\begin{tabular}[t]{  l r r r r r }
 \toprule
 Layer & \# filters & K & S   & Output  \\  
 \midrule
input  & 60 &   - &  -          & $T\times 128 $  \\  
conv1    & 64 & (3) & (1)   & $T \times 128 $  \\  
conv2 & 80 & (3)   & (1)    & $T \times 128 $  \\ 
mp2   & - & (2)   & (2)     & $T / 2 \times 128 $  \\ 
conv2 & 96 & (3)   & (1)    & $T / 2 \times 128 $  \\ 
conv2 & 112 & (3)   & (1)   & $T / 2 \times 128 $  \\ 
conv2 & 128 & (3)   & (1)   & $T / 2 \times 128 $  \\ 
attention6   & - & (1)   & (1)      & $1 \times 128 $  \\
fc7  & 128  & (1)   & (1)   & $1 \times 128$  \\
 \bottomrule
\end{tabular}

\vspace{10pt}
\caption*{{\bf (a)} Behavioral Encoder $f_b$}
\begin{tabular}[t]{  l r r r r r }
 \toprule
 Layer & \# filters & K & S   & Output  \\  
 \midrule
input  & 60 &   - &  -          & $T\times 60 $  \\  
conv1    & 64 & (3) & (1)   & $T \times 64 $  \\  
conv2 & 80 & (3)   & (1)    & $T \times 80 $  \\ 
mp2   & - & (2)   & (2)     & $T / 2 \times 80 $  \\ 
conv2 & 96 & (3)   & (1)    & $T / 2 \times 96 $  \\ 
conv2 & 112 & (3)   & (1)   & $T / 2 \times 112 $  \\ 
conv2 & 128 & (3)   & (1)   & $T / 2 \times 128 $  \\ 
attention6   & - & (1)   & (1)      & $1 \times 128 $  \\
fc7  & 128  & (1)   & (1)   & $1 \times 128$  \\
 \bottomrule
\end{tabular}

\end{center}
\normalsize
\normalsize

\vspace{7pt}
\caption{\textbf{Architecture details.}  Shown are half of the neural encoder $f_n$ and behavior encoder $f_b$ functions. How these encoders are used is shown in  \textbf{Fig.~{\color{pearDark}3}}. Neural encoder $f_n$ is followed by 1D convolutions similar to the behavioral encoder $f_b$, by replacing the number of filters. Both encoders produce $128$ dimensional output, while first half of the neural encoder do not downsample on the temporal axis. \textit{mp} denotes a max-pooling layer. Batch Normalization and ReLU activation are added after every convolutional layer. }

\label{table:encoder}
\end{table}

\parag{BehaveNet \cite{behavenet}:} This uses a discrete autoregressive hidden Markov model (ARHMM) to decompose 3D motion information into discrete “behavioral syllables." As in the regression baseline, the neural information is used to predict the posterior probability of observing each discrete syllable. Unlike the original method, we used 3D poses instead of RGB videos as targets. We skipped compressing the behavioral data using a convolutional autoencoder because, unlike RGB videos, 3D poses are already low-dimensional.

\parag{SimCLR \cite{simclr}:} We trained the original SimCLR module without the calcium imaging data and swapping augmentations. As in our approach, we took the features before the projection layer as the final representation. 

\parag{Gradient Reversal Layer (GRL) \cite{GRL}:} Together with the contrastive loss, we trained a two-layer MLP domain discriminator per modality, $D_{b}$ and $D_{n}$, which estimates the domain of the neural and behavioral representations. Discriminators were trained by minimizing
\begin{equation}
\begin{array}{r}
\mathcal{L}_{D}=\sum_{x \in\{\mathbf{b}, \mathbf{n}\}}-d \log \left(D_{m}\left(f_{m}(x)\right)\right) \; \\
\end{array}
\end{equation}
where $d$ is the one-hot identity vector. Gradient Reversal layer is inserted before the projection layer. Given the reversed gradients, the neural and behavioral encoders $f_n$ and $f_{b}$ learn to fool the discriminator and outputs invariant representations across domains, hence acting as a domain adaptation module. We kept the hyperparameters of the discriminator the same as in previous work \cite{munro20multi}. We froze the weights of the discriminator for the first 10 epochs, and trained only the $\mathcal{L}_{NCE}$. We trained the network using both loss functions, $\mathcal{L}_{NCE} + \lambda_{D} \mathcal{L}_{D}$, for the remainder of training. We set the hyperparameters $\lambda_{D}$ to $10$ empirically.

\paragraph{Maximum Mean Discrepancy (MMD) \cite{MMD}:} We replaced adversarial loss in  GRL  baseline with a statistical test that minimizes the distributional discrepancy from different domains.
 
 \paragraph{MM-SADA \cite{munro20multi}:} A recent multi-modal domain adaptation model for action recognition that minimizes cross-entropy loss on target labels, adverserial loss for domain adaptation, and contrastive losses to maximize consistency between multiple modalities. As we do not assume any action labels during the contrastive training phase, we removed the cross-entropy loss.

 \paragraph{SeqCLR \cite{eegcontrastive}:} This approach learns a uni-modal self-supervised contrastive model. Hence, we only apply it to  the neural imaging data, without using the behavioral modality. As this method was previously applied on datasets with Electroencephalography (ECoG) imaging technique, we removed ECoG specific augmentations.  \looseness-1

\paragraph{Maximum Mean Discrepancy (MMD):} We replaced adversarial loss in GRL baseline with a statistical test to minimize the distributional discrepancy from different domains \cite{MMD}. Similar to previous work, we applied MMD only on the representations before the projection layer independently on both modalities \cite{munro20multi, kangcontrastive}. Similar to the GLR baseline, we first trained 10 epochs only using the contrastive loss, and trained using the combined losses  $\mathcal{L}_{NCE} + \lambda_{MMD} \mathcal{L}_{MMD}$ for the remainder. We set the hyperparameters $\lambda_{MMD}$ as $1$ empirically.
For the domain adaptation methods GRL and MMD, we reformulated the denominator of the contrastive loss function. Given a domain function $dom$ which gives the domain of the data sample, we replaced one side of $L_{NCE}$ in Eq.~\ref{eq:nce} with,

\begin{equation}
\log \frac{\exp \left(\left\langle\mathbf{z}^{i}_{b}, \mathbf{z}^{i}_{n}\right\rangle / \tau\right)}{\sum_{k=1}^{N}  \mathbf{1}_{[dom(i) = dom(k)]}  \exp \left(\left\langle\mathbf{z}^{i}_{b}, \mathbf{z}^{k}_{n}\right\rangle / \tau\right)},
\end{equation}

where selective negative sampling prevents the formation of trivial negative pairs across domains, therefore making it easier to merge multiple domains. Negative pairs formed during contrastive learning try to push away inter-domain pairs, whereas domain adaptation methods try to merge multiple domains to close the domain gap. We found that the training of contrastive and domain adaptation losses together could be quite unstable, unless the above changes were made to the contrastive loss function.

\begin{figure}[t]
  \centering
  \includegraphics[width=.45\textwidth]{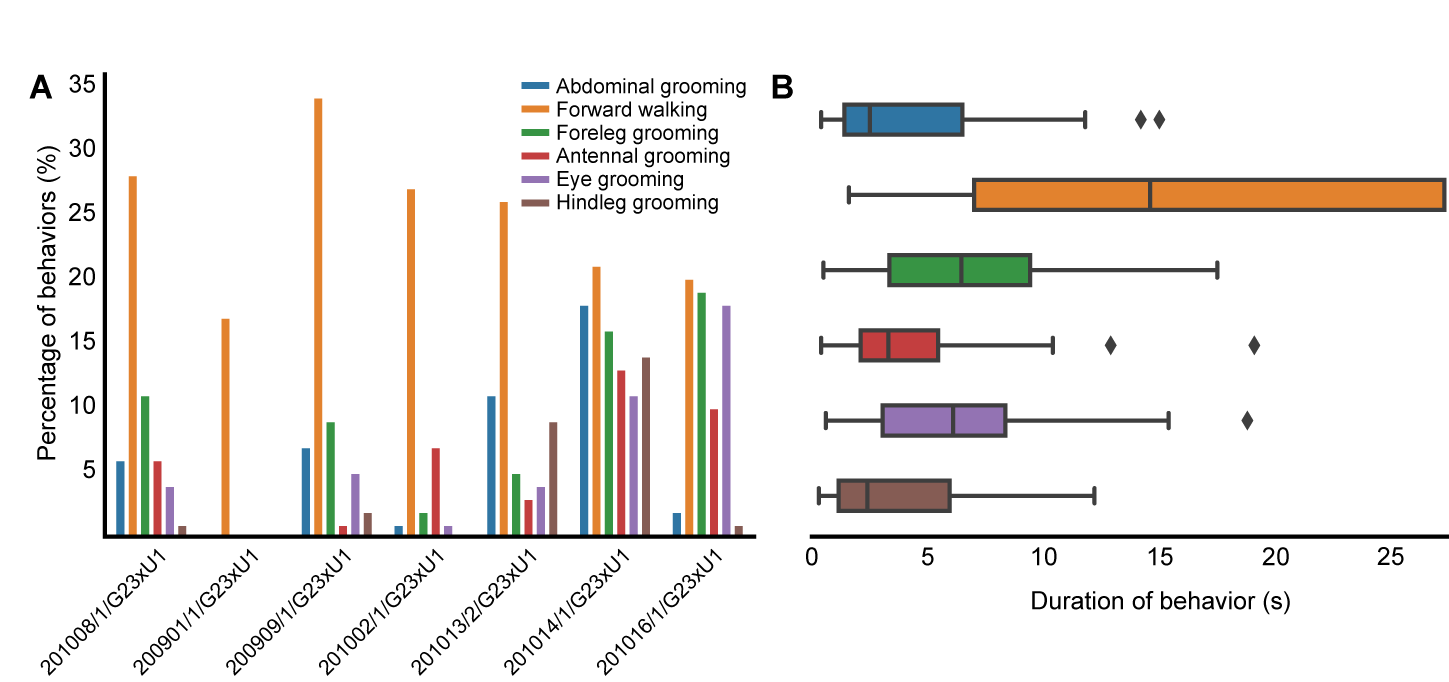}
  \caption{\textbf{Motion Capture and two-photon dataset statistics.} Visualizing \textbf{(A)} the number of annotations per animal and \textbf{(B)} the distribution of the durations of each behavior across animals. Unlike scripted human behaviors, animal behaviors occur spontaneously. The total number of behaviors and their durations do not follow a uniform distribution, therefore making it harder to model.}
  \label{fig:dataset}
\end{figure}

 \begin{figure*}[t]
  \centering
  \includegraphics[width=1\textwidth]{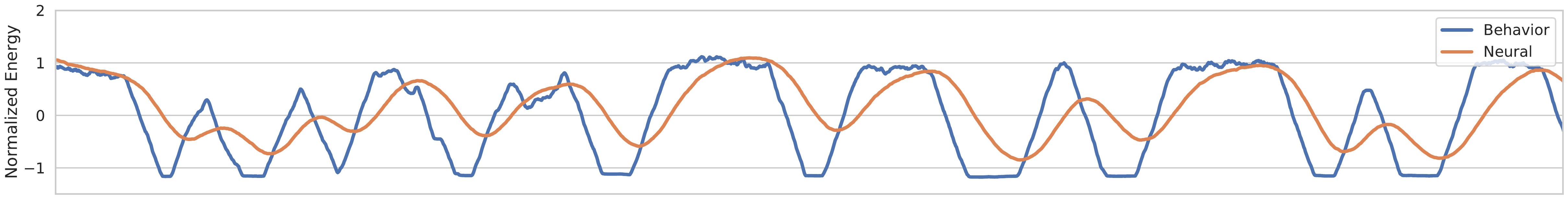}
   \includegraphics[width=1\textwidth]{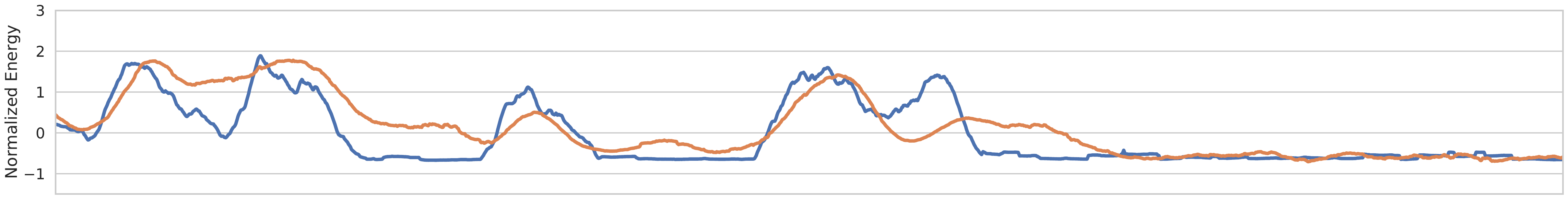}
   \includegraphics[width=1\textwidth]{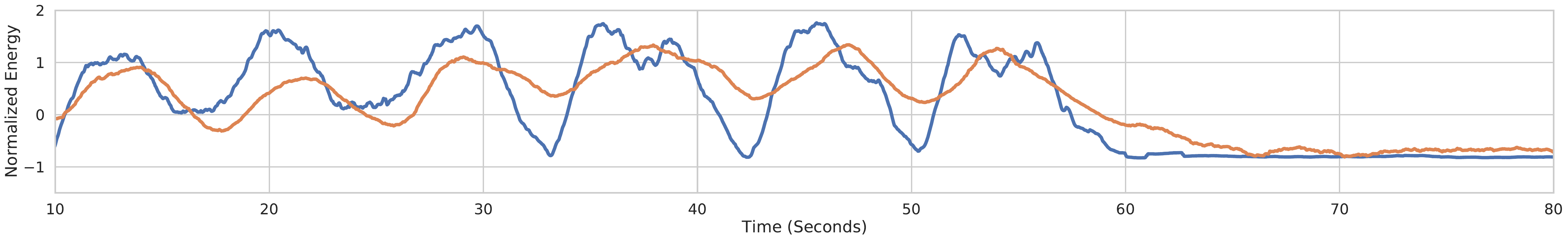}
  \caption{\textbf{Visualizing the temporal correlation between behavioral and neural energies on multiple animals.} The behavioral and neural energies are calculated as the normalized distances between consecutive frames. The multi-modal energies show a similar temporal pattern. The slower neural energy decay is due to the calcium dynamics.}
  \label{fig:energy}
\end{figure*}

\newpage
\section{Supplementary Videos}
\label{sup:video}
\paragraph{Motion Capture and Two-Photon (MC2P) Dataset.} The following videos are sample behavioral-neural recordings from two different flies. The videos show \textbf{(left)} raw behavioral RGB video together with \textbf{(right)} registered and denoised neural images at their original resolutions. The behavioral video is resampled and synchronized with the neural data. The colorbar indicates normalized relative intensity values. Calculation of $\Delta F / F$  is previously explained under Dataset Collection section.   \\

\noindent\textbf{Video 1:} \url{https://drive.google.com/file/d/1Cepy5xjLj4XiQUITY_yKKu2B4WKdl6nx}

\noindent\textbf{Video 2:}
\url{https://drive.google.com/file/d/1OSszc_fMR2Ol2WkUdj1E4u58rFVaMr6E}

\paragraph{Action Label Annotations.} Sample behavioral recordings from multiple animals using a single camera. Shown are eight different action labels: \textit{forward walking}, \textit{pushing}, \textit{hindleg grooming}, \textit{abdominal grooming}, \textit{foreleg grooming}, \textit{antennal grooming}, \textit{eye grooming} and \textit{resting}. Videos are temporally down-sampled. Animals and labels are randomly sampled. \\

\noindent\textbf{Video 3:}
\url{https://drive.google.com/file/d/1cnwRRyDZ4crrVVxRBbx32Za-vlxSP7sy}

\paragraph{Animal Motion Capture.} Sample behavioral recordings with 2D poses from six different camera views. Each color denotes a different limb. The videos are temporally down-sampled for easier view. \\

\noindent\textbf{Video 4:}
\url{https://drive.google.com/file/d/1uYcL7_Zl-N0mlG1VTrg67s2Cy71wml5S}

\noindent\textbf{Video 5:}
\url{https://drive.google.com/file/d/1eMcP-Ec1c4yBQpC4CNv45py7gObmuUeA}

\end{document}